\documentclass[final,5p,times,twocolumn,authoryear]{elsarticle}

\usepackage{amssymb}
\usepackage{amsthm}
\usepackage{subcaption}
\usepackage{hyperref}
\usepackage{microtype}
\usepackage{graphicx}
\usepackage{booktabs} 
\usepackage[table,dvipsnames]{xcolor}
\usepackage{pifont}

\usepackage{amsmath}
\usepackage{mathtools}
\usepackage{makecell}
\usepackage{multirow}
\usepackage{csquotes}

\journal{Elsevier}

\begin{document}

\begin{frontmatter}



\title{Mapping Hidden Heritage: Self-supervised Pre-training on High-Resolution LiDAR DEM Derivatives for Archaeological Stone Wall Detection}

\author[inst1]{Zexian Huang\corref{cor1}}
\ead{zexian.huang1@unimelb.edu.au}

\author[inst1]{Mashnoon Islam}
\ead{mashnoon.islam1@unimelb.edu.au}

\author[inst1]{Brian Armstrong}
\ead{brian.armstrong@unimelb.edu.au}

\author [inst2]{Billy Bell}
\ead{bill@gunditjmirring.com}

\author[inst1]{Kourosh Khoshelham}
\ead{k.khoshelham@unimelb.edu.au}

\author[inst1]{Martin Tomko}
\ead{tomkom@unimelb.edu.au}

\cortext[cor1]{Corresponding author}

\affiliation[inst1]{organization={The University of Melbourne},
            city={Parkville},
            postcode={3010}, 
            state={Victoria},
            country={Australia}}
            
\affiliation[inst2]{organization={Gunditj Mirring Traditional Owners Corporation},
            addressline={248 Condah Estate Road}, 
             city={Breakaway Creek},
             postcode={3303}, 
             state={Victoria},
             country={Australia}}

\begin{abstract}
Historic dry-stone walls hold significant cultural and environmental importance, serving as historical markers and contributing to ecosystem preservation and wildfire management during dry seasons in Australia. However, many of these stone structures in remote or vegetated landscapes remain undocumented due to limited accessibility and the high cost of manual mapping. Deep learning–based segmentation offers a scalable approach for automated mapping of such features, but challenges remain: 1.the visual occlusion of low-lying dry-stone walls by dense vegetation and 2.the scarcity of labeled training data. This study presents \textbf{DINO-CV}, a self-supervised cross-view pre-training framework based on knowledge distillation, designed for accurate and data-efficient mapping of dry-stone walls using \textbf{Digital Elevation Models (DEMs)} derived from high-resolution airborne LiDAR. By learning invariant geometric and geomorphic features across DEM-derived views, (i.e., Multi-directional Hillshade and Visualization for Archaeological Topography), DINO-CV addresses the occlusion by vegetation and data scarcity challenges. Applied to the \textbf{Budj Bim Cultural Landscape} at Victoria, Australia, a UNESCO World Heritage site, the approach achieves a mean Intersection over Union (\textit{mIoU}) of \textit{68.6\%} on test areas and maintains \textit{63.8\%} mIoU when fine-tuned with only 10\% labeled data. These results demonstrate the potential of self-supervised learning on high-resolution DEM derivatives for large-scale, automated mapping of cultural heritage features in complex and vegetated environments. Beyond archaeology, this approach offers a scalable solution for environmental monitoring and heritage preservation across inaccessible or environmentally sensitive regions.
\end{abstract}




\end{frontmatter}


\section{Introduction}
\label{introduction}

Dry-stone walls are significant cultural heritage structures that have been constructed for thousands of years in various regions around the world \citep{unesco2018art}. These walls are built without mortar, using carefully selected and placed stones to create durable and functional barriers \citep{unesco2018art,doneus2022confronting}. In Australia, dry-stone walls are often associated with European colonial history and are typically used for agricultural purposes, such as livestock containment and land management \citep{marshall2018australian}. They also hold cultural significance for Indigenous communities, who have their own traditional stone structures and practices \citep{jimenez2021dry}. The mapping and maintenance of dry-stone walls are essential for preserving ecosystems \citep{preti2018dry,camera2018quantifying} and supporting wildfire control during dry seasons in Australia. However, a substantial portion of stone walls located in wild Indigenous landscapes remain unidentified and undocumented due to limited accessibility and the high cost of conventional mapping approaches.

Traditional methods of mapping dry-stone walls, such as manual surveys or aerial photography, can be impractical and labor-intensive, especially in remote or rugged terrain. Recent advancements in computer vision and deep learning have enabled the development of automated, scalable mapping techniques that can efficiently detect dry-stone walls from high-resolution imagery and LiDAR data \citep{trotter2022machine,suh2023mapping,wang2024archaeological}. Supervised deep neural networks, such as ConvNets (CNNs) \citep{he2016deep}, vision transformers (ViTs) \citep{dosovitskiy2021an}, and U-Net architectures \citep{ronneberger2015u}, can be trained on annotated datasets to learn the visual and spatial features of dry-stone walls, including their shape, texture, and spatial patterns. These deep learning-based mapping approaches substantially reduce the time and cost of mapping and offer the potential for large-scale registration of undocumented structures. However, the performance of these deep supervised models heavily depends on the availability of large, labeled datasets for model training, which are often difficult and costly to obtain. Another critical challenge is the visual occlusion of low-lying stone wall structures by dense vegetation in wild landscapes, which makes it difficult for models to accurately detect and localize these features.

Self-supervised learning (SSL) has emerged as a promising approach to mitigate the challenge of limited labeled data for deep supervised models
\citep{caron2020unsupervised,caron2021emerging,chen2020simple,grill2020bootstrap,he2020momentum}. SSL methods are capable of learning object-centric visual representations from large unlabeled datasets, enabling the transfer of knowledge to downstream tasks such as image segmentation and classification. By pre-training models on large unlabeled datasets, SSL can improve the performance of supervised models trained with limited labeled data. In the context of dry-stone wall mapping, SSL is particularly beneficial for learning semantic and geometric features of stone walls from high-resolution aerial imagery or digital elevation models (DEMs) without extensive manual annotation. This is especially relevant in heritage landscapes, where the availability of labeled data may be limited due to the complexity and variability of the structures. 

To overcome the challenge of visual occlusion of low-lying stone walls by dense vegetation, the use of high-resolution DEMs and their derivatives (i.e., multi-directional hillshadings (MHS) and Visualization for Archaeological Topography (VAT)) for stone wall mapping can provide rich topographic information that enhances the visibility of terrain features in complex, vegetated environments. These DEM derivatives are typically processed from airborne LiDAR data, which can penetrate dense vegetation canopies to capture detailed ground surface information. In contrast, image-based data often struggle to generate accurate DEMs in vegetated locations where the terrain is heavily obscured \citep{white2013utility}.

Grounded in these hypotheses, this study poses the following research question: \textit{How can self-supervised learning on high-resolution DEM derivatives improve the mapping performance of low-lying dry-stone walls in complex, vegetated environments?} To answer this question, we explore the potential of self-supervised learning on high-resolution DEM derivatives for automated mapping of dry-stone walls in complex, vegetated environments. Hence, we propose a self-supervised cross-view pre-training strategy, DINO-CV, based on knowledge distillation to learn visual and structural features from multiple DEM derivatives generated from high-resolution aerial LiDAR data. The mapping performance of our approach is evaluated on the Budj Bim cultural landscape in Victoria, Australia, a UNESCO World Heritage site known for its extensive colonial dry-stone walls and significant Indigenous heritage. We demonstrate that these self-supervised features enable accurate and data-efficient mapping of stone wall structures on various computer vision backbones, including ResNet, Wide ResNet, and Vision Transformers (ViTs), through limited supervised fine-tuning. The proposed self-supervised pre-training strategy, DINO-CV, can learn structural and visual features from different terrain data views (i.e., MHS and VAT), thereby substantially reducing the need for labeled data in supervised fine-tuning. These findings highlight the potential of self-supervised learning on DEM derivatives for large-scale, automated mapping of stone walls in complex, vegetated, and data-scarce heritage environments.

\section{Background}

\subsection{DEM Visualization of Archaeological Landscapes}
Detecting archaeological features using high-resolution topographic data has become a central focus in archaeological remote sensing \citep{chase2017lidar,vstular2021airborne}. Airborne LiDAR systems provide highly accurate 3D measurements of the landscape by emitting laser pulses that penetrate vegetation, capturing the underlying bare-earth surface. This capability is especially advantageous in densely vegetated regions, where traditional aerial imagery may obscure subtle topographic variations.

These LiDAR-derived measurements can be processed into digital elevation models (DEMs), which represent the elevation of the bare-earth surface by filtering out vegetation and non-ground elements. DEMs can provide detailed topographic information that enables the detection of subtle landscape features, including ancient structures, earthworks, and other archaeological remains. As a result, DEMs have proven particularly suitable for mapping and analyzing large, complex, and vegetated archaeological landscapes \citep{vstular2012visualization}. To further enhance the visibility of these features, a variety of DEM visualization techniques have been developed, such as hillshading \citep{yoeli1967mechanisation}, slope analysis \citep{doneus2006full}, and sky-view factor \citep{kokalj2011application}. These methods aid researchers in identifying and interpreting landscape features that are often not discernible in standard aerial or satellite imagery.

One of the DEM visualization methods, Visualization for Archaeological Topography (VAT) developed by \citet{verbovvsek2019vat,kokalj2019not}, integrates multiple topographic layers, including hillshaded relief, slope, positive openness, and sky-view factor, into a single composite image. This combination method enhances the detection of subtle topographic variations that may indicate buried or obscured archaeological features. Another widely used technique, multi-directional hillshading (MHS) \citep{devereux2008visualisation}, generates multiple hillshade images from different illumination angles and combines them into a multi-band raster or composite image. This method improves the visibility of terrain features such as slopes, ridges, and embankments, while reducing shadow saturation and mitigating directional bias in DEMs. For instance, \citet{comer2019airborne} employed grayscale MHS visualizations derived from LiDAR data to reveal subtle cultivation features, including irrigation berms and channels, on Temwen Island and the islets of Nan Madol.

Together, these visualization techniques serve as powerful tools for archaeological mapping and interpretation, enabling researchers to identify, map, and analyze archaeological features embedded in complex cultural landscapes that might otherwise remain undetected \citep{guyot2021combined,guyot2021objective}. A recent study by \citet{jaturapitpornchai2024impact} evaluated the impact of various DEM visualizations on the segmentation performance of archaeological features using deep learning models. The study concludes that identifying a universally optimal visualization method remains inconclusive, as performance varies across archaeological classes and landscape contexts.

\subsection{Automated Mapping of Dry-Stone Wall Structures}
Recent advances in deep learning have enabled the automated detection and mapping of archaeological features from remote sensing data. Previous study by \citet{trotter2022machine} employed a supervised Fully convolutional Neural Network (FCN) \citep{long2015fully} based on the U-Net architecture \citep{ronneberger2015u} to automate the mapping of dry-stone walls in Denmark. Their model was trained on a dataset of digital terrain models (DTMs) with a spatial resolution of 0.4m, using ground-truth annotations of dry-stone wall locations. While the study demonstrated the potential of deep learning for this task, its applicability is limited by the availability of high-resolution terrain data, which is widely accessible in Denmark but not available in other regions. Furthermore, the Danish DTM dataset is updated only every five years, and the performance of the supervised model is constrained by the limited quantity of labeled training data.

Another study by \citet{suh2023mapping} proposed an automated mapping framework based on U-Net \citep{ronneberger2015u} and ResU-Net \citep{diakogiannis2020resunet} architectures for detecting dry-stone walls in forested terrain in the Northeastern United States. Their work investigated the impact of the quality of high-resolution DEM derivatives, specifically hillshade and slope maps, generated from airborne LiDAR data on model performance. They found that illumination parameters, such as sunlight angle and the brightness or darkness of hillshade images, were important factors that affect the model's performance on detecting small-scale geomorphic features. These findings highlight the importance of optimizing DEM visualization inputs when applying deep learning models to archaeological mapping tasks.

A related study by \citet{wang2024archaeological} developed a segmentation framework for mapping the ancient city walls from airborne LiDAR data in archaeological landscapes. Focusing on the ancient capital of Jinancheng, their method involved converting airborne LiDAR point clouds into DEMs, followed by the application of a U-Net architecture to segment wall remains at the pixel level. The study ensures the quality of labeled data of city wall by expert interpretation based on the archaeological survey data, and increase the number of training samples by data augmentation for model training. A post-processing step using connected component analysis was applied to refine the segmentation outputs of model by removing noise and filling small gaps.

Beyond archaeological applications, related research on drainage ditch mapping demonstrates the broader potential of combining deep learning with airborne laser scanning (ALS) for detecting subtle geomorphic features in vegetated environments. \citet{lidberg2023mapping} developed a deep neural network trained on ALS data and manually digitized ditch channels across ten regions in Sweden, achieving an 86\% detection accuracy. Their study highlights how deep learning models can effectively identify fine-scale hydrological and geomorphic structures that are otherwise difficult to observe due to canopy cover or terrain complexity. This approach parallels the objectives of our work on dry-stone wall mapping, emphasizing the utility of high-resolution topographic data and learning-based feature extraction for large-scale, automated landscape analysis.

These studies together highlight the potential of deep learning for the automated mapping of archaeological stone wall structures. However, a common limitation across existing approaches is the dependence on large quantities of high-quality labeled data to achieve strong segmentation performance. The generation of such datasets often requires expert knowledge and significant manual effort, which can be resource-intensive. Moreover, the limited availability of high-resolution terrain data in many regions restricts the generalizability and scalability of these methods to diverse archaeological contexts.

\subsection{Self-supervision with Knowledge Distillation}
Self-supervised learning (SSL) methods based on knowledge distillation have been proposed as an effective approach for learning object-centric visual features from large, unlabeled datasets, enabling improved performance on downstream tasks such as image segmentation and classification \citep{grill2020bootstrap,caron2021emerging}. By pre-training models on large, unlabeled datasets, SSL can improve the performance of supervised models even with limited labeled data. 

\citet{hinton2015distilling} introduced knowledge distillation as a method to transfer knowledge from a large ensemble of models (teachers) into a single, compact student model. This approach enables efficient deployment while retaining the performance benefits of model ensembles. Recent SSL methods based on knowledge distillation, such as DINO \citep{caron2021emerging} and BYOL \citep{grill2020bootstrap}, have adapted this teacher–student framework to a self-supervised, label-free setting by using momentum encoders and self-distillation to match intermediate feature representations and output distributions between teacher and student networks. These methods do not require labeled data and are well-suited for learning semantic and structural features from visual inputs. Recent extensions, DINOv2 \citep{oquab2024dinov} and DINOv3 \citep{simeoni2025dinov3}, push this paradigm to much larger scales. DINOv2 stabilizes training at scale and introduces an automated pipeline for curating large and diverse image datasets. DINOv3 further scales the approach to 7B parameters and trains on even larger curated datasets, emphasizing scalability and generalization. While recent extensions such as DINOv2 and DINOv3 push self-distillation toward foundation-scale models through large curated datasets and massive ViTs, these directions are beyond the scope of this work. Instead, our focus is on cross-view self-supervision tailored to high-resolution DEM derivatives, addressing archaeological mapping challenges where labeled data is scarce.

In parallel with self-distillation SSL approaches, contrastive self-supervised learning methods such as SimCLR \citep{chen2020simple, chen2020big}, MoCo \citep{he2020momentum}, and MoCov2 \citep{chen2020improved} learn visual features by maximizing agreement between different augmented views of the same image (positive pairs), while minimizing agreement between views of different images (negative pairs). This is typically achieved by contrasting representations in a latent embedding space. Although contrastive learning has shown strong performance across a range of visual tasks, it often requires large batch sizes or memory banks to provide a sufficient number of negative examples for effective training \citep{caron2021emerging}.

Both contrastive and distillation-based SSL frameworks have demonstrated state-of-the-art performance on various vision tasks and are particularly promising for applications in remote sensing and archaeological feature mapping, where labeled data is limited but large volumes of unlabeled imagery or elevation data are available. In the domain of remote sensing, the SeCo framework proposed by \citet{manas2021seasonal} leverages seasonal variations in satellite imagery, along with temporal and spatial invariance constraints to learn robust and temporal-invariant visual representations through contrastive learning. The knowledge distillation framework, DINO-MM \citep{wang2022self}, extends the DINO framework to multimodal data, enabling the learning of joint representations from different modalities (i.e., optical and SAR imagery). This approach has demonstrated strong performance in remote sensing tasks such as image classification and retrieval. DINO-MC \citep{wanyan2024extending} introduces a multi-size cropping strategy to strengthen global–local view alignment between teacher and student networks. By incorporating local crops of varied sizes, the model learns more diverse and scale-invariant visual features on the SeCo-100K \citep{manas2021seasonal} dataset, which enhances its performance across a range of downstream tasks, including remote sensing imagery classification and change detection.

In contrast, we propose a self-supervised cross-view pre-training framework based on knowledge distillation paradigm to pre-train vision backbones on unlabeled DEM derivatives (i.e., MHS and VAT), which can improve stone wall mapping capability in complex and vegetated archaeological landscapes.

\section{Methodology}
\label{method}

\begin{figure*}[!tbp]
\centering
\includegraphics[width=0.74\textwidth]{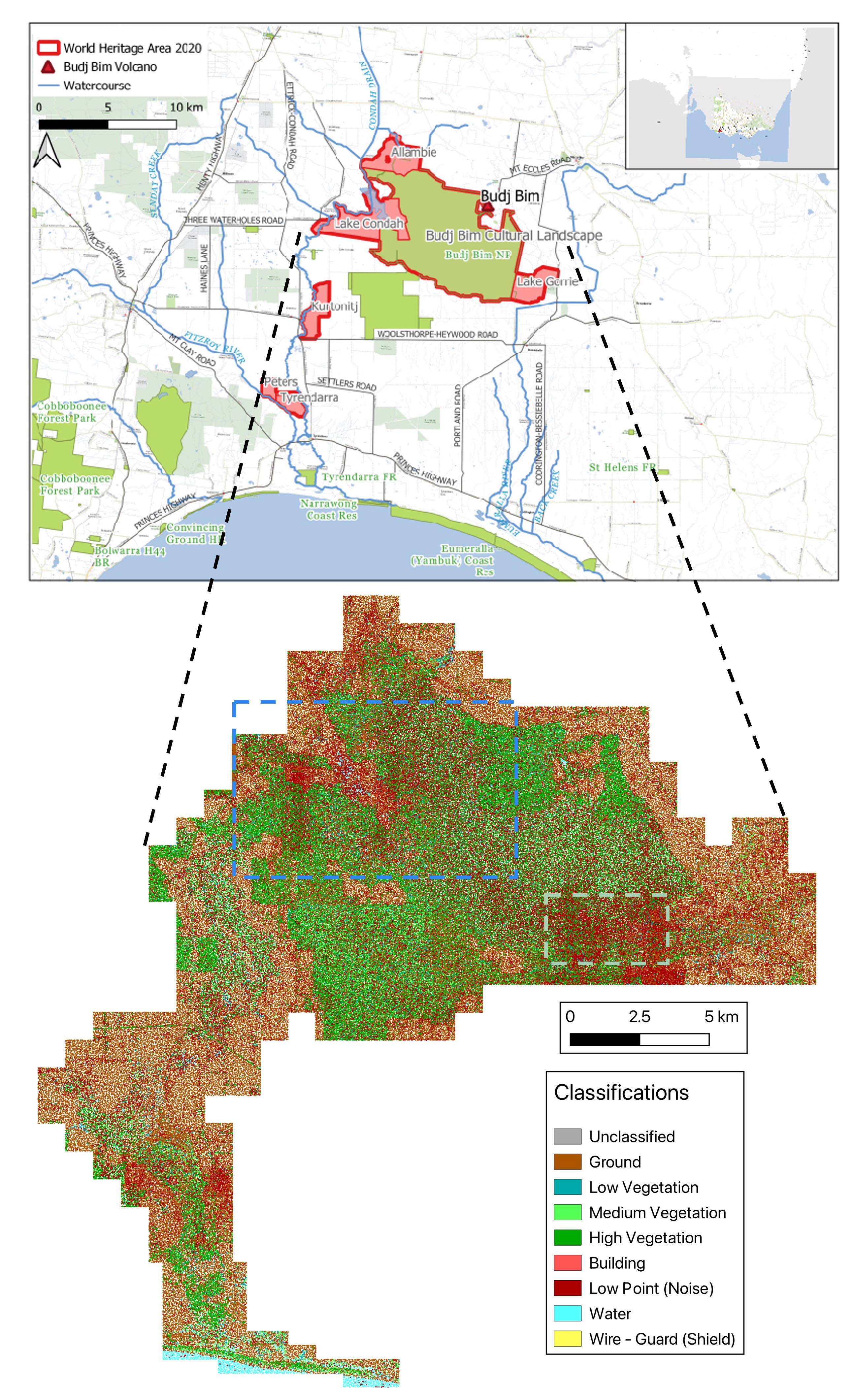}
\caption{Overview of the airborne LiDAR dataset covering 301~km\textsuperscript{2} of the Budj Bim Cultural Landscape in southwestern Victoria, Australia. The survey captured over 33~billion points using multi-return laser scanning, providing the high-resolution topographic data required for generating digital elevation models (DEMs) and derivative visualizations used in this study. The location of \textit{BudjbimArea (25~km\textsuperscript{2})} dataset for fine-tuning is shown in a blue dashed bounding box, and the location of an unlabeled region used for qualitative demonstration is shown in a green dashed bounding box.}
\label{fig:budjbim_bev}
\end{figure*}

\begin{figure*}[!tbp]
\centering
\includegraphics[width=1\textwidth]{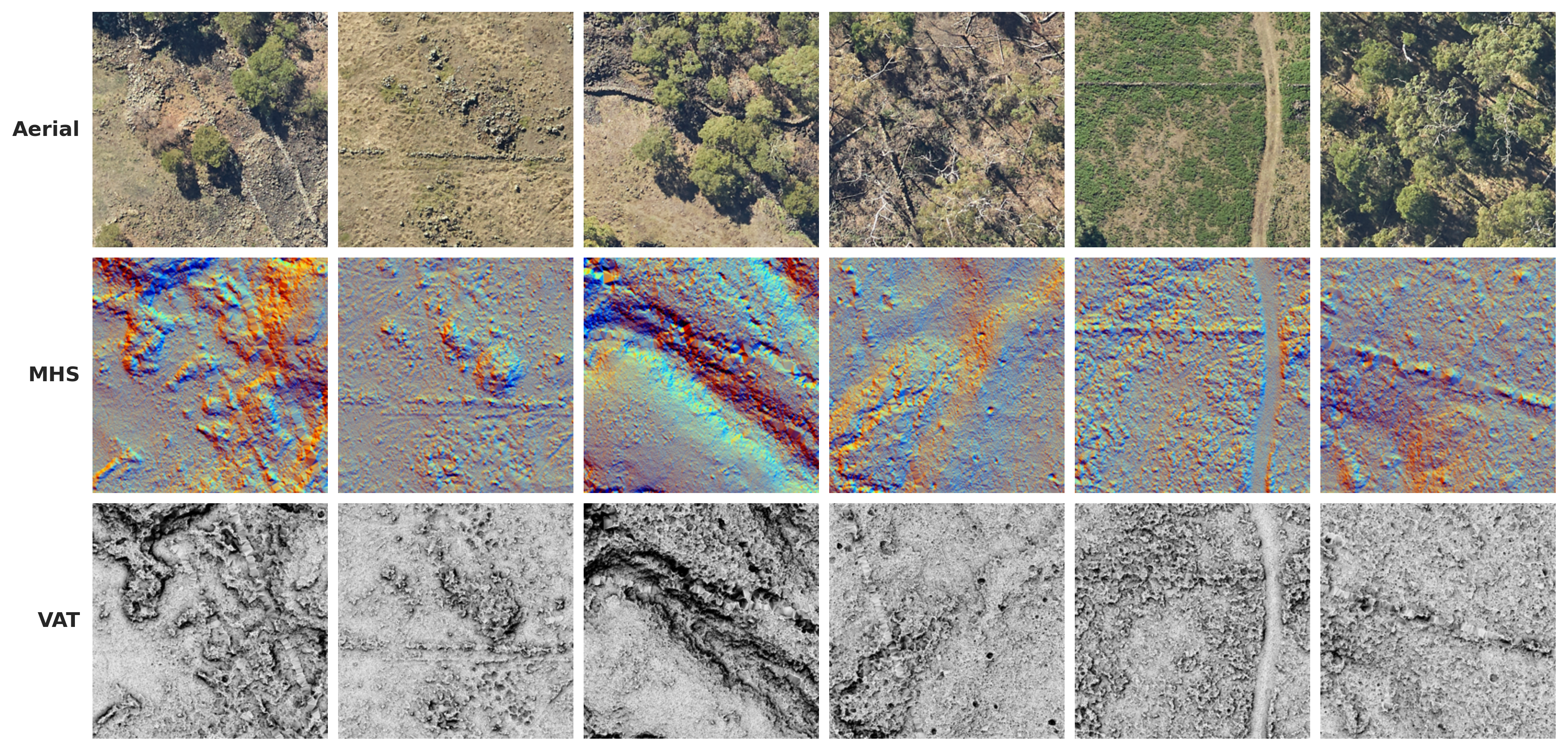}
\caption{The aerial image (Aerial, Row~1), Multi-directional Hillshade (MHS, Row~2), and Visualization for Archaeological Topography (VAT, Row~3) of dry-stone walls in the Budj Bim landscape. From left to right, columns show: (1-2) multiple parallel thin walls and an isolated thin wall; (3-4) thick, zigzag walls and a straight wall partially occluded by vegetation; (5-6) roadside straight walls with gate openings, and a heavily occluded wall segment.}
\label{fig:aerial_mhs_vat}
\end{figure*}

\subsection{Airborne LiDAR Acquisition \& Processing}
\label{subsec:lidar_processing}

To enable reliable detection of subtle archaeological features across the Budj Bim landscape, careful consideration was given to the timing and conditions of airborne LiDAR acquisition. The LiDAR survey was conducted in 2020 by the Department of Environment, Land, Water and Planning (DELWP), Victoria, on behalf of the Gunditjmara Traditional Owners, during the dry season and following bushfire events, when vegetation cover was minimal. These conditions maximized ground visibility and enhanced the detectability of subtle surface features. Capturing fine-scale archaeological traces such as dry-stone walls required high point densities, achieved through multiple overlapping flight passes at varying scan angles to ensure uniform surface coverage across the diverse terrain (Fig.~\ref{fig:budjbim_bev}).

Following data acquisition, the raw LiDAR point cloud was processed to isolate ground points from vegetation and other non-terrain elements. The Cloth Simulation Filtering (CSF) algorithm \citep{zhang2016easy} was adopted for this purpose due to its robustness in modelling complex ground surfaces under variable vegetation conditions. The CSF algorithm simulates a virtual cloth draped over an inverted point clouds, effectively distinguishing ground points by conforming to the underlying terrain surface while excluding elevated canopy structures.

Ground-classified points were subsequently interpolated to generate high-resolution DEMs at 10~cm spatial resolution. Rasterization and gridding were applied with carefully selected interpolation parameters to ensure smooth and accurate terrain representation. The resulting raster DEMs preserve the geomorphic characteristics of the landscape, including low-relief archaeological features that are often imperceptible in optical imagery. These processed DEMs served as the foundational layers for generating the DEM derivatives and visualizations used in the DINO-CV pre-training framework.

\subsection{DEM Visualization of the Budj Bim Landscape}
\label{subsec:dem_vis}

High-resolution DEM visualization techniques are useful visualization tools for mapping cultural heritage features in complex terrains, where archaeological structures such as dry-stone walls are often obscured in conventional aerial imagery. To enhance the visibility of low-lying structures and support automated stone-wall mapping on archaeological landscapes, the high-resolution DEMs are processed using Relief Visualization Toolbox (RVT) \citep{kokalj2011application,kokalj2019not} to generate two DEM-derived visualization methods: Multi-directional Hillshade (MHS) and Visualization for Archaeological Topography (VAT).

MHS simulates illumination from multiple sun angles to enhance the perception of topographic features, such as ridges, edges, and linear structures embedded in the landscape. This visualization technique provides a general-purpose view of terrain geometry, improving the contrast and continuity of linear archaeological features. In contrast, VAT emphasizes micro-topographic variations by combining multiple elevation derivatives, including slope, local relief, openness, and sky-view factor, into a single composite image. VAT is designed specifically to highlight subtle archaeological features such as buried walls, ditches, and mounds that may not be visible in traditional hillshading.

While aerial imagery provides high-fidelity semantic information, such as textures of objects and surrounding vegetation, it is often limited by occlusions caused by dense vegetation, particularly in rural or forested areas. In contrast, MHS and VAT visualizations derived from DEMs are unaffected by such occlusions and can effectively reveal dry-stone wall structures that are otherwise indiscernible in aerial views, as illustrated in Fig.~\ref{fig:aerial_mhs_vat}. In detail, columns 1 and 2 of Fig.~\ref{fig:aerial_mhs_vat} highlight multiple thin, parallel stone walls and a solitary linear segment, which are clearly discernible in both MHS and VAT. Columns 3 and 4 display zigzag and straight thick walls that are partially or fully concealed by vegetation in the aerial view but remain visible in the DEM visualizations. Columns 5 and 6 show roadside walls with gate openings and heavily occluded segments.

Together, MHS and VAT offer complementary topographic features that are well-suited for detecting the diverse forms of dry-stone walls across the Budj Bim cultural landscape. Their integration enables robust feature extraction and significantly enhances automated mapping performance where visual occlusion and subtle relief variations pose challenges to traditional image-based methods.

\subsection{Self-supervised Cross-view Pre-training}

\begin{figure*}[!tbp]
\centering
\includegraphics[width=1\textwidth]{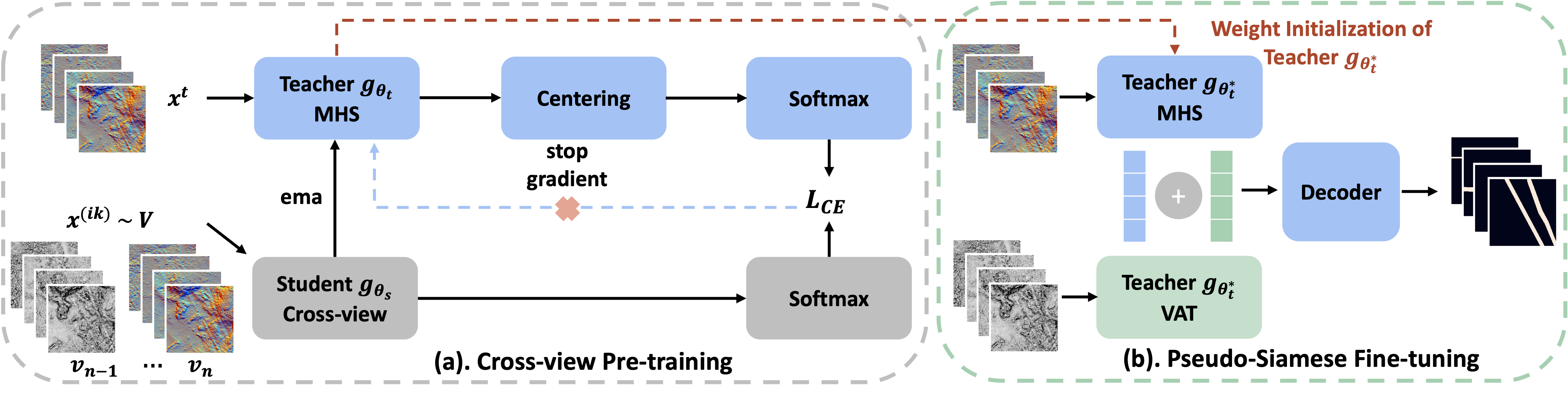}
\caption{\textbf{Overview of the DINO-CV architecture.} \textbf{(a) Self-supervised Cross-view Pre-training:} The framework employs a student-teacher distillation setup. At each iteration $k$, the student network $g_{\theta_s}$ receives a randomly sampled input view $x^{(ik)}$ (e.g., VAT) from the collection $\mathcal{V}$, while the teacher network $g_{\theta_t}$ receives a fixed view $x^{t}$ (e.g., MHS). The student is optimized via stochastic gradient descent to match the teacher's output, while the teacher parameters are updated via an \textbf{Exponential Moving Average (ema)} of the student weights. \textbf{(b) Pseudo-Siamese Fine-tuning:} For the downstream task, we instantiate a Pseudo-Siamese two-stream network initialized with weights from two separate teacher networks from DINO-CV pre-training (one specialized on MHS, one on VAT). These parallel encoders extract complementary geometric features which are fused via element-wise summation before being passed to the decoder for stone wall segmentation.}
\label{fig:dino_cv_arch}
\end{figure*}

\paragraph{Knowledge Distillation} 
The self-supervised cross-view pre-training framework, DINO-CV (Fig.~\ref{fig:dino_cv_arch}a), is designed to learn invariant visual and structural features across multiple high-resolution DEM derivatives to enhance mapping performance of models in the subsequent supervised fine-tuning. DINO-CV builds on the knowledge distillation paradigm introduced by \citet{caron2021emerging}, where a student network $g_{\theta_s}$, comprising a vision backbone $f_{\theta_s}$ and a projection head $h_{\theta_s}$, such that $g_{\theta_s} = h_{\theta_s} \circ f_{\theta_s}$, is trained to match the output distribution of a teacher network $g_{\theta_t}$ by minimizing the cross-entropy between their temperature-scaled softmax outputs:
\begin{equation}
    \min_{\theta_s} \; \mathcal{L}\left(P_t(g_{\theta_t}(x)), P_s(g_{\theta_s}(x))\right),
    \label{eq:cross_entropy_loss}
\end{equation}
where $P_t$ and $P_s$ are the softmax-normalized outputs of the teacher and student networks, respectively.

During training, the student parameters $\theta_s$ are updated using stochastic gradient descent (SGD) to minimize the distillation loss. The teacher network, however, is not directly optimized; instead, its parameters $\theta_t$ are updated at each iteration using an exponential moving average (EMA) of the student’s parameters \citep{tarvainen2017mean}:
\begin{equation}
    \theta_t \leftarrow m \theta_t + (1 - m) \theta_s,
    \label{eq:ema_update}
\end{equation}
where $m \in [0, 1)$ is a momentum coefficient that controls the smoothness of the teacher update.

To prevent model collapse, the output of the teacher network is centered by adding a running mean vector $c$ before softmax normalization:
\[
g_{\theta_t}(x) \leftarrow g_{\theta_t}(x) + c.
\]
The center $c$ is updated at each iteration based on the current batch of teacher outputs:
\begin{equation}
    c \leftarrow m c + (1 - m) \frac{1}{B} \sum_{i=1}^{B} g_{\theta_t}(x_i),
    \label{eq:center_update}
\end{equation}
where $B$ is the batch size. In addition, output sharpening is applied by using a low temperature $\tau_t$ in the teacher softmax, which concentrates the probability mass and encourages the student to produce confident predictions.

\paragraph{Cross-view Pre-training} 
The core idea of DINO-CV is learning invariant visual and structural terrain features across multiple DEM derivatives (e.g., MHS and VAT). In the cross-view pre-training strategy, the student network is trained to learn view-invariant features by matching stable targets derived from varying data inputs. Let $\mathcal{V} = \{v_1,...,v_n\}$ be a collection of DEM derivatives (e.g., MHS and VAT), and let $x^{(i)} \sim \mathcal{V}$ be a randomly sampled data view for the student, while $x^{(t)}$ denotes a fixed view (e.g., MHS) for the teacher.

Following the multi-crop augmentation strategy of \citet{caron2021emerging}, multiple global and local crops are generated from both inputs: $v_s = \text{aug}(x^{(i)})$ and $v_t = \text{aug}(x^{(t)})$. These crops are passed into the networks to compute a set of representations from the student: $\{g_{\theta_s}(v^{c}_s)\}_{c=1}^{N}$, where $N$ is the number of student crops (global + local), and a set of global representations from the teacher: $\{g_{\theta_t}(v^{c}_t)\}_{c=1}^{M}$, where $M$ is the number of teacher global crops. The student is trained to match the teacher’s output distributions by minimizing the cross-entropy loss in Eq.~\ref{eq:cross_entropy_loss} across all teacher-student crop pairs.

\paragraph{Convergence of the Teacher Network} 
In the pre-training of DINO-CV, while the teacher network $g_{\theta_t}$ receives only a fixed data view (e.g., MHS or VAT) as input, the EMA update (Eq.~\ref{eq:ema_update}) ensures that the teacher progressively inherits the learned cross-view invariance and semantically meaningful representations from the student network $g_{\theta_s}$, as follows:
\begin{equation}
\theta_t^{(k+1)} = m \theta_t^{(k)} + (1 - m) \theta_s^{(k+1)},
\label{eq:ema_update_teacher}
\end{equation}
where $m \in [0,1)$ is a momentum coefficient controlling the smoothness of parameters update.

At each iteration $k$, the student parameters $\theta_s^{(k)}$ are updated by SGD with a randomly sampled data view $x^{(i_k)} \sim \mathcal{V}$ from the collection of DEM derivatives:
\begin{equation}
\theta_s^{(k+1)} = \theta_s^{(k)} - \eta \nabla_{\theta_s} \mathcal{L}_k,
\label{eq:sgd_update_student}
\end{equation}
where $\eta$ is the learning rate for weight update, and $\mathcal{L}_k$ is the cross-entropy loss between the teacher’s output on the fixed view $x^{(t)}$ and the student’s output on the random view $x^{(i_k)}$. This however introduces non-stationarity into the training process, as the student’s input view distribution changes over time.

Despite this randomness in student inputs, the teacher can still converge, assuming that the distillation loss $\mathcal{L}_k$ consistently aligns the student’s outputs to stable targets. Since the teacher's output is anchored to a fixed input data view and the student is repeatedly trained to match this output under varying inputs, the student gradually learns to produce view-invariant representations. Consequently, the teacher accumulates and smooths these representations via the EMA of student weights.

By unrolling the EMA update from Eq.~\ref{eq:ema_update_teacher}, the teacher’s parameters update at iteration $k$ can be expressed as:
\begin{equation}
\theta_t^{(k)} = (1 - m) \sum_{i=1}^{k} m^{k - i} \theta_s^{(i)} + m^k \theta_t^{(0)},
\label{eq:ema_update_teacher_unroll}
\end{equation}
which is a weighted moving average of the past student parameters.

Assuming that the student parameters converge to a stable point:
\[
\lim_{k \to \infty} \theta_s^{(k)} = \theta^*,
\]
as expected when the loss $\mathcal{L}_k$ decreases and the randomly varying input data views are semantically aligned, the teacher parameters also converge:
\[
\lim_{k \to \infty} \theta_t^{(k)} = \theta^*.
\]
This ensures that the teacher network generates stable, smooth, and semantically consistent targets throughout the cross-view training process.

\subsection{Pseudo-Siamese Network}
In supervised fine-tuning, we adopt a \textit{Pseudo-Siamese Network} architecture (Fig.~\ref{fig:dino_cv_arch}b) to effectively leverage the complementary representations of DEM derivatives. This architecture consists of two parallel teacher networks pre-trained with DINO-CV, each trained on a different data view. A \textit{Pseudo-Siamese Network} refers to a two-stream architecture where the networks share the same architecture but have \textit{independent weights}. In our case, the MHS and VAT streams are parameterized by their respective teacher networks, each pre-trained separately with DINO-CV, allowing the two streams to specialize in their respective data view.

Let $g_{\theta_t^{\text{MHS}}}$ and $g_{\theta_t^{\text{VAT}}}$ denote the two pre-trained teacher networks: the teacher network trained with MHS inputs $g_{\theta_t^{\text{MHS}}}(x^{\text{MHS}})$ and with VAT inputs $g_{\theta_t^{\text{VAT}}}(x^{\text{VAT}})$.

During supervised fine-tuning, each data-view-specific input is passed into its corresponding teacher network, and their latent representations are fused via element-wise summation:
\begin{equation}
z(x) = g_{\theta_t^{\text{MHS}}}(x^{\text{MHS}}) + g_{\theta_t^{\text{VAT}}}(x^{\text{VAT}}).
\label{eq:psn_fusion}
\end{equation}

The fused representation $z(x)$ is then passed to a decoder head $Dec(z(x))$ for pixel-wise semantic segmentation of dry-stone walls. Depending on the encoder backbone, we use: 1. U-Net decoder \citep{ronneberger2015u} for ConvNet backbones (e.g., ResNet-50 and Wide ResNet-50), and 2. DPT decoder \citep{ranftl2021vision} for ViT backbones (e.g., ViT-S/16). This architectural design also benefits from the view-invariant representations learned by the student network during cross-view pre-training. Since the parameters of teacher networks are derived from the student networks trained to align outputs across randomly sampled data views, the fused representations $z(x)$ encode robust, view-invariant structural features transferable across different data views.

\subsection{Implementation}
\label{subsec:impl}

\paragraph{Pre-training}
We pre-train three widely used vision backbones: ResNet-50 \citep{he2016deep}, Wide-ResNet-50-2 \citep{zagoruyko2016wide}, and ViT-S/16 \citep{dosovitskiy2021an}, using the DINO-CV framework. The choice of ResNet-50 and ViT-S/16 is motivated by their comparable model sizes (21M vs. 22M parameters) and contrasting architectural paradigms (convolutional vs. attention-based), enabling a fair comparison of different vision backbones for the dry-stone wall mapping. Wide-ResNet-50-2 is selected for its increased representational capacity through a wider architecture, which has demonstrated strong performance across diverse vision tasks.

Pre-training follows the established implementation protocols of DINO \citep{caron2021emerging}. ViT-S/16 is optimized using AdamW \citep{loshchilov2018decoupled}, while ResNet-50 and Wide-ResNet-50-2 are trained using SGD with momentum. All models are trained with a batch size of 256 for 100 epochs, distributed over two NVIDIA A100 GPUs. The learning rate is linearly warmed up during the first 10 epochs to an initial value of 0.0005 for ViT-S/16 and 0.001 for the ResNet variants, followed by a cosine decay schedule. The base momentum parameter $m$ for the EMA update (Eq.~\ref{eq:ema_update}) of the teacher network is initialized at 0.996 and gradually increased to 1.0 following a cosine schedule.

We apply data augmentations consistent with DINO \citep{caron2021emerging} and BYOL \citep{grill2020bootstrap}, including the multi-crop strategy with both global and local views. For ViT-S/16, global crops are randomly sampled from scale ranges $(0.33, 1.0)$ and local crops from $(0.1, 0.33)$. For ResNet-based models, global crops are randomly sampled from $(0.15, 1.0)$ and local crops from $(0.05, 0.15)$.

\paragraph{Fine-tuning}
We fine-tune the pre-trained vision backbones for 100 epochs on the labeled BudjBimArea dataset (Sec.~\ref{subsec:dataset}) for stone wall segmentation. Hyperparameters for fine-tuning are selected based on a combination of prior work on DINO \citep{caron2021emerging} and empirical validation for training stability under limited labeled data. To maintain compatibility with the learned representations, we retain the same optimizers used during self-supervised pre-training (AdamW for ViT-S/16 and SGD for ResNet-based backbones).

ResNet-based models are optimized using SGD with a batch size of 64, an initial learning rate of 0.01, and a weight decay of 0.0001. For ViT-S/16, we employ the AdamW optimizer with a batch size of 64, a learning rate of 0.00005, and a weight decay of 0.05. The reduced batch size compared to pre-training reflects the increased memory footprint of the pseudo-Siamese architecture (Fig.~\ref{fig:dino_cv_arch}b), which processes MHS and VAT inputs in parallel. For ViT-S/16, a small learning rate is used to preserve the geometric and semantic representations learned during self-supervised pre-training, as larger learning rates were observed to cause unstable convergence and feature degradation. In contrast, ResNet-based models optimized with SGD require a comparatively larger learning rate to achieve effective fine-tuning.

For data augmentation in fine-tuning, input data are resized to $256 \times 256$ using bicubic interpolation, followed by random rotations up to 180 degrees. Image normalization is performed according to ImageNet statistics \citep{deng2009imagenet}. For mask inputs, resizing and rotations are performed using nearest-neighbor interpolation to preserve label integrity.

\section{Experiments}
\label{experiments}

\begin{figure*}[!tbp]
\centering
\includegraphics[width=1\textwidth]{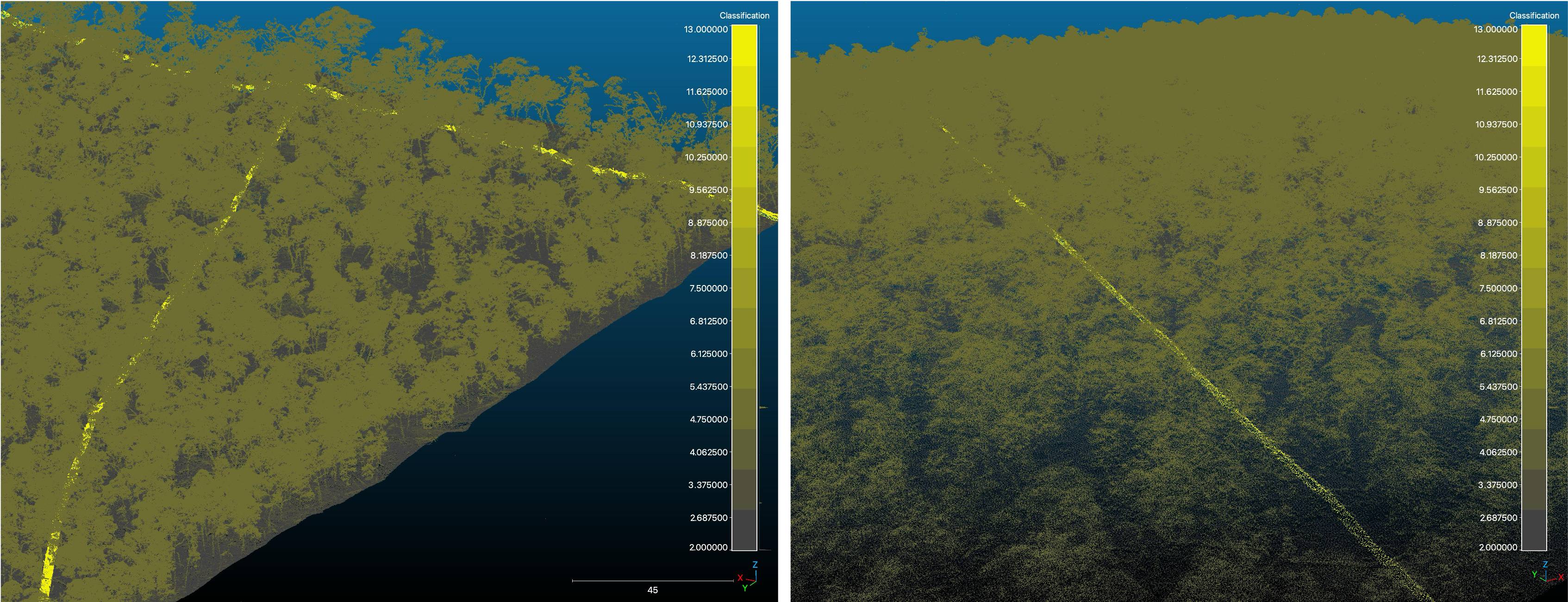}
\caption{Oblique 3D visualization of the annotated airborne LiDAR point clouds within the \textit{BudjBimArea} dataset. The dry-stone wall structures are highlighted in light yellow (Classification ID 13), demonstrating their visibility beneath the dense vegetation canopy. Despite significant occlusion from trees and undergrowth, the high density of the LiDAR survey allows for the continuous delineation of these low-relief features. These walls typically measure approximately 0.5-1.0~m in width and 0.5-1.2~m in height, as derived from the ground point measurements.}
\label{fig:bba_pc_anno}
\end{figure*}

The experimental setting in this study evaluates \textit{intra-site generalization} (i.e., spatial generalization within a single landscape) under a self-supervised learning paradigm. Self-supervised pre-training is conducted on the \textit{BudjBimLandscape} dataset using only unlabeled DEM-derived inputs, while supervised fine-tuning and evaluation are restricted to labeled samples from the spatially distinct \textit{BudjBimArea} subset. This setting reflects a common and practically relevant scenario in archaeological remote sensing, where extensive LiDAR coverage of a site is available but only a limited fraction can be annotated. The objective is therefore to maximize label efficiency and intra-site generalization within a single landscape.

\subsection{Dataset}
\label{subsec:dataset}

Table~\ref{tab:datasets} provides a comprehensive summary of the key characteristics of the datasets used in this study. 
\paragraph{BudjBimLandscape}  
For self-supervised pre-training, we use the \textit{BudjBimLandscape} dataset, which covers approximately 301~km\textsuperscript{2} of terrain within the Budj Bim Cultural Landscape in southeastern Australia. The dataset comprises unlabeled aerial imagery and two DEM-derived visualization products: MHS and VAT, both generated from a high-resolution LiDAR-derived DEM at 10~cm spatial resolution (see Sec.~\ref{subsec:dem_vis}). Due to the substantial visual occlusion of low-lying stone walls by dense vegetation in optical imagery, only the MHS and VAT representations are used during DINO-CV self-supervised pre-training. The dataset contains a total of 188,006 image tiles of size $400 \times 400$ pixels, which are resized to $256 \times 256$ for training.

\begin{figure*}[!tbp]
\centering
\includegraphics[width=1\textwidth]{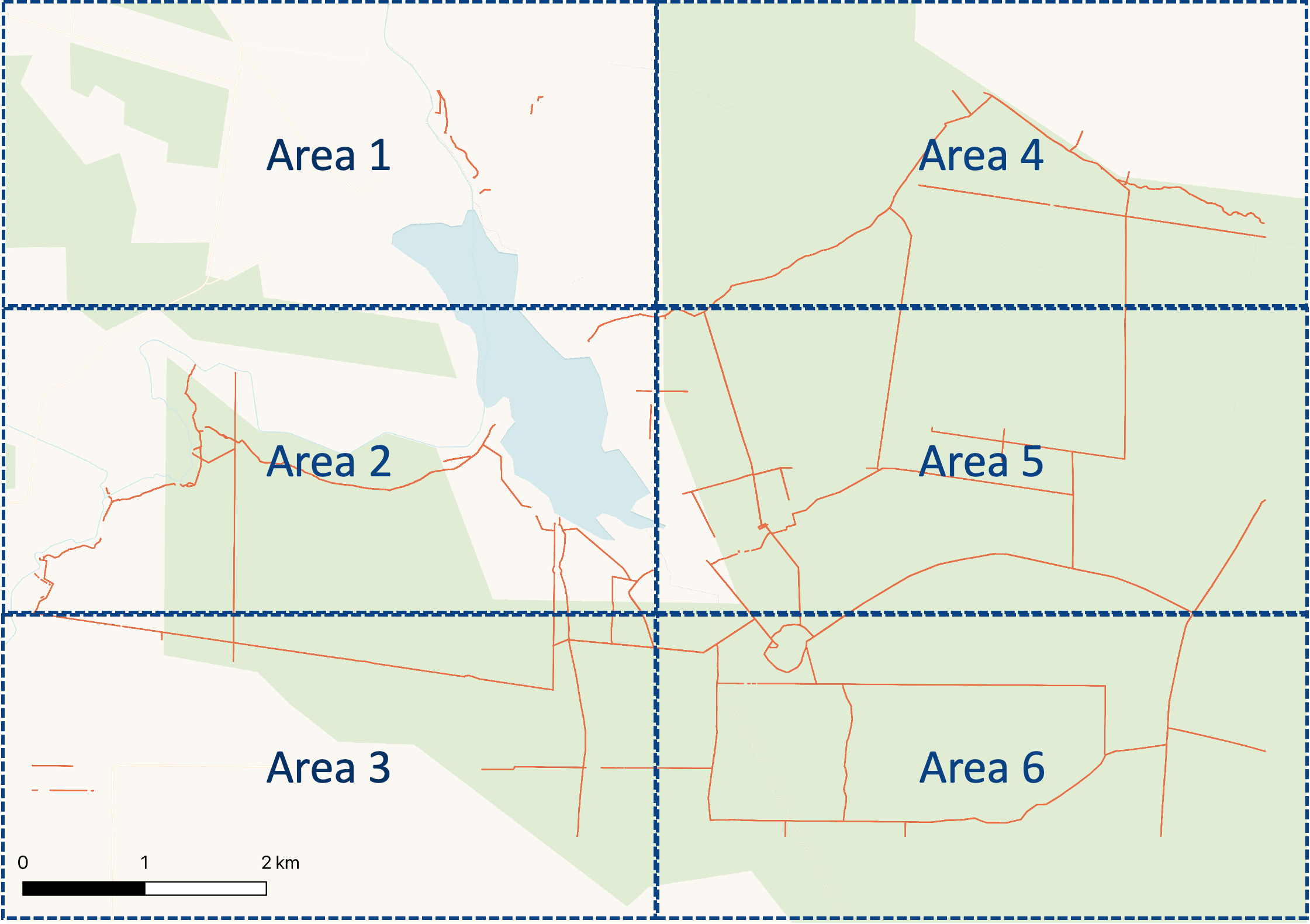}
\caption{Spatial partitioning of the \textit{BudjBimArea} dataset. Orange lines indicate annotated European historic dry-stone walls near Tae Rak (Lake Condah), Victoria, Australia. The number of data samples per area is as follows: Area 1 - 186, Area 2 - 1128, Area 3 - 1138, Area 4 - 1233, Area 5 - 1604, and Area 6 - 1813.}
\label{fig:bba_dataset}
\end{figure*}

\paragraph{BudjBimArea}  
For supervised fine-tuning and evaluation, we use the annotated \textit{BudjBimArea} dataset, a geographical subset of the larger \textit{BudjBimLandscape}, covering a 25~km\textsuperscript{2} region in the northern part of the Budj Bim landscape. This area contains one of the highest densities of historic European dry-stone walls in Australia. The region is spatially partitioned into six non-overlapping rectangular zones, as illustrated in Fig.~\ref{fig:bba_dataset}. The dataset comprises 7,102 image tiles, each with a resolution of $400 \times 400$ pixels, resized to $256 \times 256$ for fine-tuning and evaluation. We adopt a six-fold \textit{leave-one-area-out} cross-validation strategy, where in each fold five areas are used for training and the remaining area is held out for testing. Within the training subset, data is further split into 80\% for training and 20\% for validation. This strategy ensures that every tile is evaluated exactly once and provides a robust assessment of \textit{intra-site generalization} across spatially distinct regions.

\paragraph{Data Annotations}
To ensure the reliability of the ground truth labels, a multi-modal annotation strategy is adopted to integrate raster-based labeling with geometric verification using LiDAR data. Dry-stone wall annotations were manually delineated as binary raster masks in the $xy$-plane using the Semi-Automatic Classification Plugin within QGIS \citep{congedo2021semi}. While individual DEM derivatives (i.e., MHS and VAT) might exhibit visual noise due to vegetation cover and terrain roughness, cross-referencing these data views with high-resolution aerial imagery (Fig.~\ref{fig:aerial_mhs_vat}) enabled the reliable identification of linear structures of dry-stone walls and their distinction from natural topographic features. To further reduce false positives associated with elevated vegetation that may overlap wall footprints in 2D raster views, we incorporated geometric verification using the underlying 3D LiDAR point clouds (Fig.\ref{fig:bba_pc_anno}). In practice, a height-above-ground filter was applied during label assignment. Height above ground was estimated using a local ground surface model derived from neighboring ground points, and points exceeding 2~m above the estimated ground surface were excluded. This threshold reflects the typical vertical extent of dry-stone walls in the study area (approximately 0.5--1.2~m in height). This annotation workflow substantially reduces high-canopy noise and improves label consistency, ensuring that the resulting binary masks accurately represent the spatial extent of archaeological stone wall features.

\begin{table*}[!tbp]
\centering
\caption{Summary of the \textit{BudjBimLandscape} and \textit{BudjBimArea} datasets used for self-supervised pre-training and supervised fine-tuning.}
\label{tab:datasets}
\begin{tabular}{lcc}
\toprule
\textbf{Description} & \cellcolor{gray!10}{\textbf{BudjBimLandscape}} & \cellcolor{LimeGreen!10}{\textbf{BudjBimArea}} \\
\midrule
Usage & \cellcolor{gray!10}{Self-supervised pre-training} & \cellcolor{LimeGreen!10}{Supervised fine-tuning and evaluation} \\
Coverage Area & \cellcolor{gray!10}{301~km\textsuperscript{2}} & \cellcolor{LimeGreen!10}{25~km\textsuperscript{2}} \\
Location & \cellcolor{gray!10}{Budj Bim Cultural Landscape, VIC} & \cellcolor{LimeGreen!10}{Northern Budj Bim region} \\
Data Views & \cellcolor{gray!10}{MHS, VAT} & \cellcolor{LimeGreen!10}{MHS, VAT, binary wall annotations} \\
DEM Resolution & \cellcolor{gray!10}{10~cm} & \cellcolor{LimeGreen!10}{10~cm} \\
Tile Size & \cellcolor{gray!10}{$400 \times 400$ px (resized to $256 \times 256$)} & \cellcolor{LimeGreen!10}{$400 \times 400$ px (resized to $256 \times 256$)} \\
Number of Tiles & \cellcolor{gray!10}{188{,}006} & \cellcolor{LimeGreen!10}{7{,}102} \\
Label Availability & \cellcolor{gray!10}{Unlabeled} & \cellcolor{LimeGreen!10}{Manual wall annotations} \\
Data Splitting & \cellcolor{gray!10}{Not applicable} & \cellcolor{LimeGreen!10}{6-fold leave-one-area-out cross-validation} \\
\bottomrule
\end{tabular}
\end{table*}

\subsection{Results}

\paragraph{Evaluation Metrics}  
For supervised evaluation, we assess model performance on dry-stone wall mapping using standard semantic segmentation metrics: Precision, Recall, F1 Score, and mean Intersection over Union (mIoU). These metrics are computed at the pixel level and averaged across individual image tiles to ensure a consistent and fine-grained evaluation. 

\paragraph{Quantitative Result}

\begin{table*}[htbp]
\small
\centering
\caption{Stone wall mapping performances of ConvNet and ViT architectures. Evaluation metrics (i.e., Precision, Recall, F1 Score, and mIoU) are averaged over individual image tiles from the \textit{BudjBimArea} dataset. The best performance for each backbone is highlighted in bold.}
\begin{tabular}{lcccccccc}
\toprule
Method & Backbone & Param. & Pre-train Dataset & Pre. & Rec. & F1 & mIoU \\
\midrule
Random Init. & ResNet-50-Siam. & 46M & - & 74.7 & 77.0 & 75.8 & 61.1 \\
Supervised & ResNet-50-Siam. & 46M & ImageNet-1K & 76.9 & 81.4 & 79.1 & 65.4 \\
DINO \citep{caron2021emerging} & ResNet-50-Siam. & 46M & ImageNet-1K  & 75.7 & 74.3 & 75.0 & 60.0 \\
DINO-MC \citep{wanyan2024extending} & ResNet-50-Siam. & 46M & SeCo-100K  & 78.0 & 80.1 & 79.0 & 65.3 \\
\rowcolor{pink!40}
DINO-CV (ours) & ResNet-50-Siam. & 46M & BudjBimLand. & \textbf{79.4} & \textbf{83.0} & \textbf{81.2} & \textbf{68.3} \\
\midrule
Random Init. & WRN-50-2-Siam. & 138M & -  & 75.6 & 77.7 & 76.6 & 62.1 \\
Supervised & WRN-50-2-Siam. & 138M & ImageNet-1K & 76.4 & 82.5 & 79.3 & 65.7 \\
DINO-MC \citep{wanyan2024extending} & WRN-50-2-Siam. & 138M & SeCo-100K  & 79.0 & 79.2 & 79.1 & 65.4 \\
\rowcolor{pink!40}
DINO-CV (ours) & WRN-50-2-Siam. & 138M & BudjBimLand. & \textbf{79.8} & \textbf{83.0} & \textbf{81.4} & \textbf{68.6} \\
\midrule
Random Init. & ViT-S/16-Siam. & 42M & - & 77.1 & 73.0 & 75.0 & 60.0 \\
Supervised & ViT-S/16-Siam. & 42M & ImageNet-1K & 74.2 & 82.4 & 78.1 & 64.0 \\
DINO \citep{caron2021emerging}& ViT-S/16-Siam. & 42M & ImageNet-1K  & 76.8 & 81.1 & 78.9 & 65.2 \\
DINO-MC \citep{wanyan2024extending} & ViT-S/8-Siam. & 42M & SeCo-100K  & \textbf{77.4} & 80.4 & 78.9 & 65.1 \\
\rowcolor{pink!40}
DINO-CV (ours) & ViT-S/16-Siam. & 42M & BudjBimLand. & \textbf{77.4} & \textbf{83.0} & \textbf{80.1} & \textbf{66.8} \\
\bottomrule
\end{tabular}
\label{tab:fine_tune}
\end{table*}

Table~\ref{tab:fine_tune} reports the supervised stone wall mapping performance on three backbone architectures: ResNet-50, Wide-ResNet-50-2, and ViT-S/16. The models are evaluated under different pre-training strategies: random weight initialization, supervised pre-training on ImageNet-1K, self-supervised pre-training on ImageNet-1K via DINO, SeCo-100K pre-training using DINO-MC, and our proposed DINO-CV on the BudjBimLandscape dataset. The results demonstrate that DINO-CV uniformly outperforms all other pre-training baselines across all model architectures. Explicitly, models pre-trained with DINO-CV achieve the highest F1 scores (ResNet50: 81.2\%, Wide ResNet50: 81.4\% and ViT-S/16: 80.1\%) and mIoU (ResNet50: 68.3\%, Wide ResNet50: 68.6\% and ViT-S/16: 66.8\%), indicating strong capability to detect and localize dry-stone walls in complex terrain. These performance gains are evident over both supervised and self-supervised baselines, highlighting the benefit of domain-specific, cross-view self-supervised pre-training enabled by DINO-CV. Importantly, the improvements hold across both ConvNet and ViT backbones, suggesting the strong model-agnostic generalization and robustness of the proposed pre-training strategy.

\begin{figure*}[!tbp]
\centering
    \subfloat[ResNet50]{\includegraphics[width=1\textwidth]{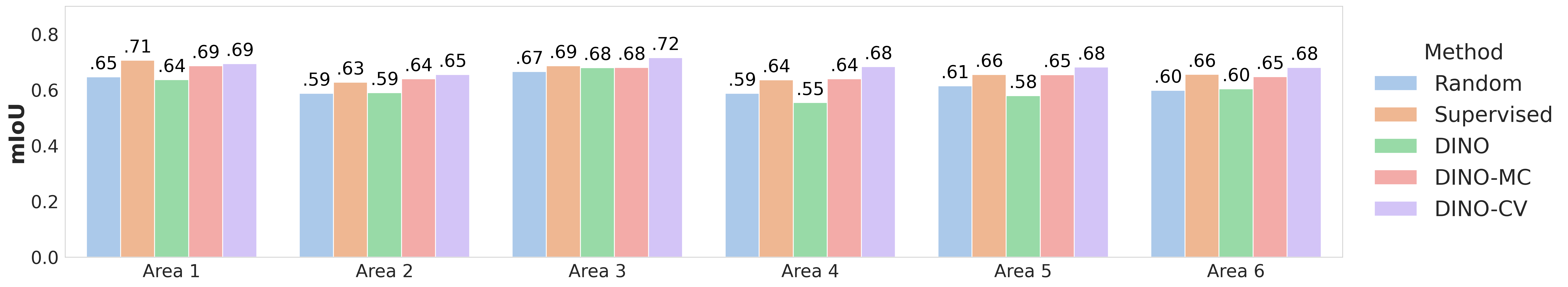}} 
    \hfill
    \subfloat[Wide ResNet50]{\includegraphics[width=1\textwidth]{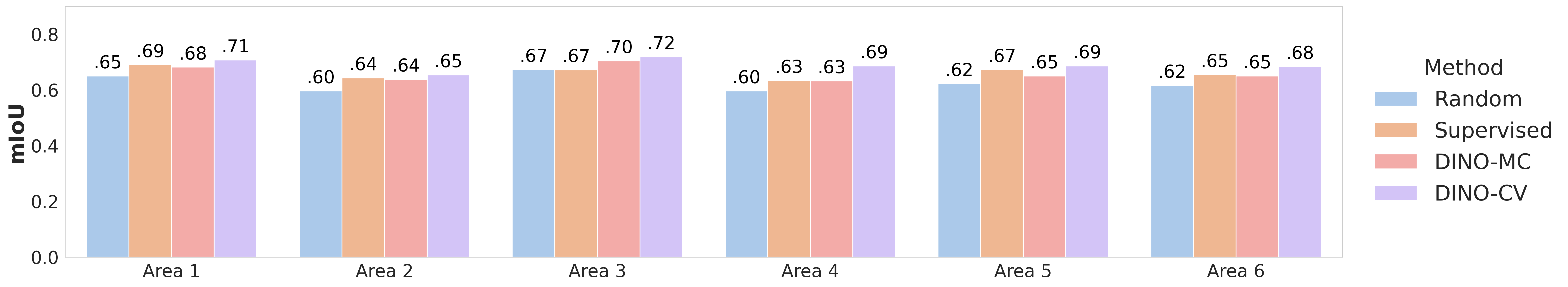}} 
    \hfill
    \subfloat[ViT-S/16]{\includegraphics[width=1\textwidth]{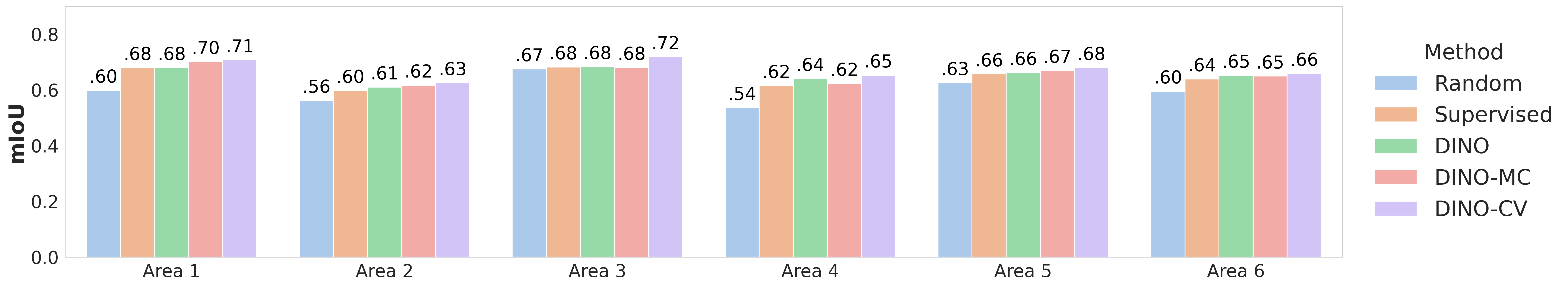}} 
\caption{Test results (mIoU\%) of each vision backbone on BudjBimArea dataset (area 1-6).}
\label{fig:results_by_areas}
\end{figure*}

Figure~\ref{fig:results_by_areas} provides a detailed comparison of mIoU scores across six areas of the BudjBimArea dataset for each model backbone and pre-training strategy. Overall, DINO-CV achieves the highest or comparable performance relative to all baseline methods. For both ResNet-50 and Wide ResNet-50-2, DINO-CV shows notable improvements in regions with challenging visual characteristics, such as dense vegetation or ambiguous terrain boundaries (e.g., Area~3 and Area~6). ViT-S/16 also benefits significantly from DINO-CV, particularly in areas where other pre-training methods struggle. For example, in Area~2 and Area~4, DINO-CV with ViT-S/16 backbone improves mIoU by 7–10 percentage points compared to random initialization. These results further support the robustness and intra-site generalization of DINO-CV across diverse spatial regions of the dataset and architectural designs.

\paragraph{Qualitative Result}

\begin{figure*}[!tbp]
\centering
\includegraphics[width=0.73\textwidth]{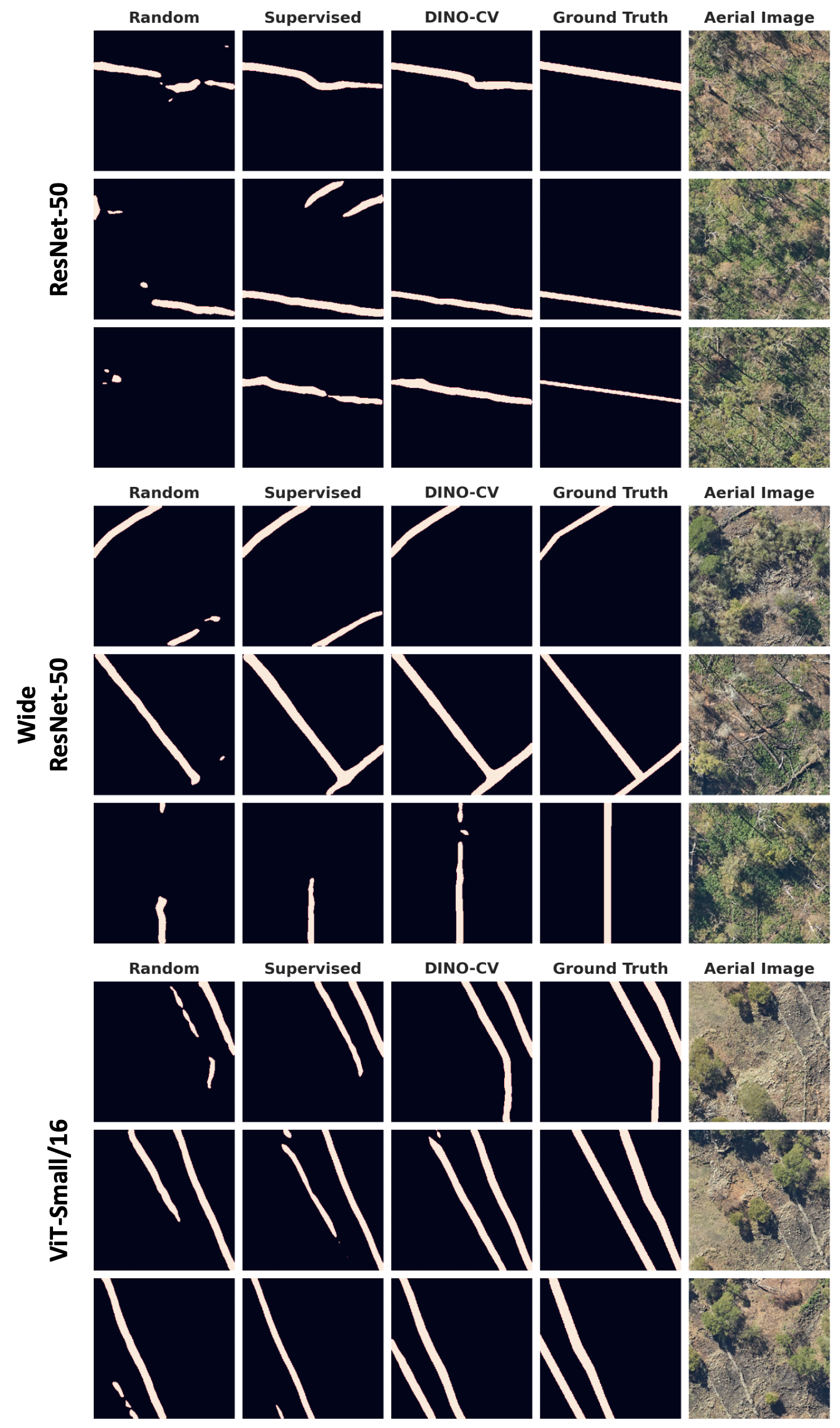}
\caption{Qualitative results of predictions using ResNet-50, Wide ResNet-50-2, and ViT-S/16 backbones under three pre-training strategies: random initialization, supervised training on ImageNet-1K, and DINO-CV (columns 1–3). Column 4 shows the ground truth (GT) segmentation masks, and column 5 provides the corresponding aerial imagery of the landscape.}
\label{fig:dino_cv_preds}
\end{figure*}

Figs.~\ref{fig:dino_cv_preds} present qualitative results of dry-stone wall mapping using the ResNet-50, Wide ResNet-50-2, and ViT-S/16 backbones under three pre-training strategies: random initialization, supervised pre-training, and DINO-CV. Each figure illustrates the predicted segmentation masks for three representative areas from the BudjBimArea dataset.

The results demonstrate that models pre-trained with DINO-CV produce more accurate and spatially coherent segmentation masks compared to those initialized randomly or pre-trained via supervised learning. In Area~3 (Fig.~\ref{fig:dino_cv_preds}, ResNet-50), the ResNet-50 model with DINO-CV accurately captures intricate wall structures and subtle boundaries, whereas the randomly initialized baseline fails to detect substantial portions of the wall segments. Similarly, in Area~4 (Fig.~\ref{fig:dino_cv_preds}, Wide ResNet50), the Wide ResNet-50-2 model with DINO-CV more effectively delineates complex wall boundaries in vegetated terrain, outperforming both the randomly initialized and supervised baselines, which misclassify several stone wall fragments. ViT-S/16 also shows notable qualitative improvements in Area~1 (Fig.~\ref{fig:dino_cv_preds}, ViT-S/16), where DINO-CV enhances the integrity of parallel wall segmentation, compared to the fragmented predictions of the other baselines. These qualitative results align with the quantitative performance trends and highlight the effectiveness of DINO-CV in improving the mapping capacity of both ConvNet and ViT backbones across diverse and challenging landscape settings.

\paragraph{Generalization to Unlabeled Region}

\begin{figure*}[!tbp]
\centering
\includegraphics[width=1\textwidth]{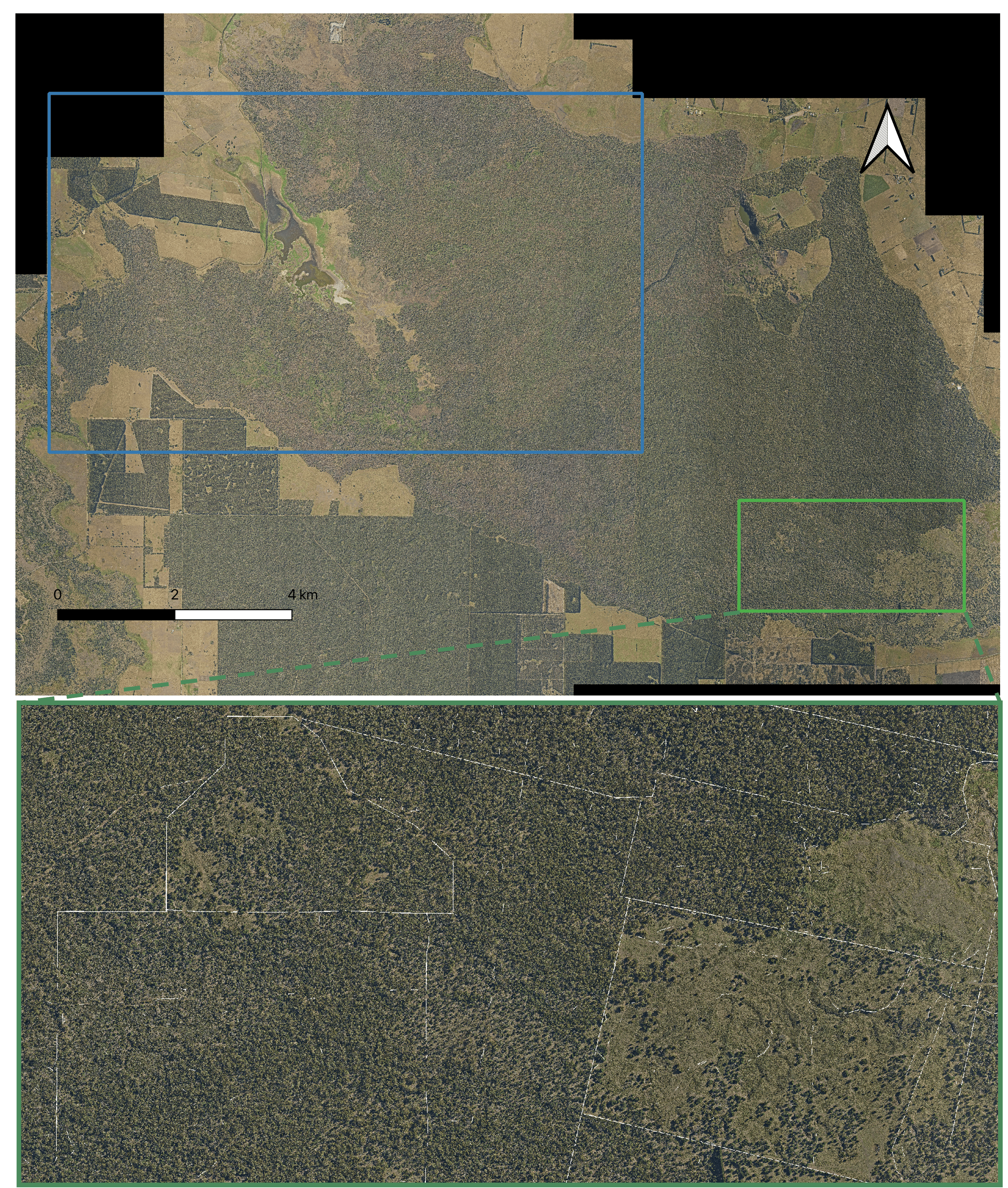}
\caption{Intra-site generalization performance of the DINO-CV model with a ResNet-50-Siamese backbone on an unlabeled region within the Budj Bim Cultural Landscape. The upper panel shows the aerial imagery, with the blue rectangle indicating the annotated training area and the green rectangle denoting the spatially separated unlabeled region. The lower panel provides a zoomed view of the unlabeled region, where the model delineates linear dry-stone wall structures (shown in white) beneath dense vegetation.}
\label{fig:unlabeled_pred_map}
\end{figure*}

To qualitatively assess the spatial generalization of DINO-CV beyond the annotated region, the fine-tuned ResNet-50-Siamese model was applied to an unlabeled area (green rectangle) located approximately 10~km from the training area (blue rectangle), as shown in Fig.~\ref{fig:unlabeled_pred_map}. Despite the absence of labeled samples, the model successfully delineates linear dry-stone wall structures (shown in white) that are largely obscured by dense vegetation. This qualitative result suggests that cross-view self-supervised pre-training on DEM derivatives (MHS and VAT) enables the model to learn transferable geomorphic and geometric latent features that remain effective across spatially separated regions within the same landscape. While this experiment does not constitute independent quantitative validation, it provides supporting evidence for the robustness and scalability of the proposed approach in unlabeled and logistically inaccessible regions.

\section{Discussion}
\label{discussion}

\begin{table*}[htbp]
\small
\centering
\caption{Ablation study on the DINO-CV. We analyze the effectiveness of two key design components:  (1) the effect of cross-view self-supervised pre-training of DINO-CV, and (2) the use of the Pseudo-Siamese Architecture for supervised fine-tuning. Each row reports the F1 score and mean Intersection over Union (mIoU) for a specific configuration, evaluated on the BudjBimArea dataset. The columns $\Delta_{F1}$ and $\Delta_{mIoU}$ denote the performance difference relative to the DINO-CV baseline (row 1, 5, and 9 for each backbone).}
\begin{tabular}{lccccccc}
\toprule
Method & Architecture & Input & Cross-view & F1 & \cellcolor{gray!40}$\Delta_{F1}$ & mIoU & \cellcolor{gray!40}$\Delta_{mIoU}$ \\
\midrule
1 DINO-CV & ResNet-50-Siam. & MHS+VAT & \checkmark & 81.2 & \cellcolor{gray!40} - & 68.3 & \cellcolor{gray!40} - \\
\textcolor{gray}{2} & ResNet-50-Siam. & MHS+VAT & \cellcolor{gray!40}{$\times$} & 80.9 & \cellcolor{gray!40}{-0.3} & 67.9 & \cellcolor{gray!40}{-0.4} \\
\textcolor{gray}{3} & ResNet-50 &\cellcolor{gray!40}{MHS} & \checkmark & 80.8 & \cellcolor{gray!40}{-0.4} & 67.8 & \cellcolor{gray!40}{-0.5} \\
\textcolor{gray}{4} & ResNet-50 & \cellcolor{gray!40}{VAT} & \checkmark & 80.9 & \cellcolor{gray!40}{-0.3} & 68.0 & \cellcolor{gray!40}{-0.3} \\
\midrule
5 DINO-CV & WRN-50-2-Siam. & MHS+VAT & \checkmark & 81.4 & \cellcolor{gray!40} - & 68.6 & \cellcolor{gray!40} - \\
\textcolor{gray}{6} & WRN-50-2-Siam. & MHS+VAT & \cellcolor{gray!40}{$\times$} & 81.0 & \cellcolor{gray!40}{-0.4} & 68.0 & \cellcolor{gray!40}{-0.6} \\
\textcolor{gray}{7} & WRN-50-2 & \cellcolor{gray!40}{MHS} & \checkmark & 80.8 & \cellcolor{gray!40}{-0.6} & 67.7 & \cellcolor{gray!40}{-0.9} \\
\textcolor{gray}{8} & WRN-50-2 & \cellcolor{gray!40}{VAT} & \checkmark & 81.0 & \cellcolor{gray!40}{-0.4} & 68.1 & \cellcolor{gray!40}{-0.5} \\
\midrule
9 DINO-CV & ViT-S/16-Siam. & MHS+VAT & \checkmark & 80.1 & \cellcolor{gray!40} - & 66.8 & \cellcolor{gray!40} - \\
\textcolor{gray}{10} & ViT-S/16-Siam. & MHS+VAT & \cellcolor{gray!40}{$\times$} & 79.6 & \cellcolor{gray!40}{-0.5} & 66.2 & \cellcolor{gray!40}{-0.6} \\
\textcolor{gray}{11} & ViT-S/16 & \cellcolor{gray!40}{MHS} & \checkmark & 79.2 & \cellcolor{gray!40}{-0.9} & 65.6 & \cellcolor{gray!40}{-1.2} \\
\textcolor{gray}{12} & ViT-S/16 & \cellcolor{gray!40}{VAT} & \checkmark & 79.4 & \cellcolor{gray!40}{-0.7} & 65.8 & \cellcolor{gray!40}{-1.0} \\
\bottomrule
\end{tabular}
\label{tab:ablation}
\end{table*}

\begin{table*}[htbp]
\small
\centering
\caption{Model mIoU (\%) Performance on BudjBimArea training subsets.}
\begin{tabular}{lcccc}
\toprule
Method & Backbone & 50\% & 30\% & 10\% \\
\midrule
Random Init. & ResNet-50-Siam. & 51.6 & 38.3 & 21.4 \\
Supervised & ResNet-50-Siam. & 64.9 & 62.9 & 49.3 \\
\rowcolor{pink!40}
DINO-CV & ResNet-50-Siam. & \textbf{67.7} & \textbf{67.2} & \textbf{63.6} \\
\midrule
Random Init. & WRN-50-2-Siam. & 53.6 & 38.4 & 22.5 \\
Supervised & WRN-50-2-Siam. & 65.3 & 63.6 & 49.7 \\
\rowcolor{pink!40}
DINO-CV & WRN-50-2-Siam. & \textbf{68.0} & \textbf{66.7} & \textbf{63.8} \\
\midrule
Random Init. & ViT-S/16-Siam. & 52.9 & 42.8 & 14.3 \\
Supervised & ViT-S/16-Siam. & 63.2 & 61.6 & 56.4 \\
\rowcolor{pink!40}
DINO-CV & ViT-S/16-Siam. & \textbf{65.1} & \textbf{63.2} & \textbf{58.9} \\
\bottomrule
\end{tabular}
\label{tab:subset}
\end{table*}

\begin{figure*}[!tbp]
\centering
    \subfloat[Attention feature maps on MHS inputs.]{\includegraphics[width=1\textwidth]{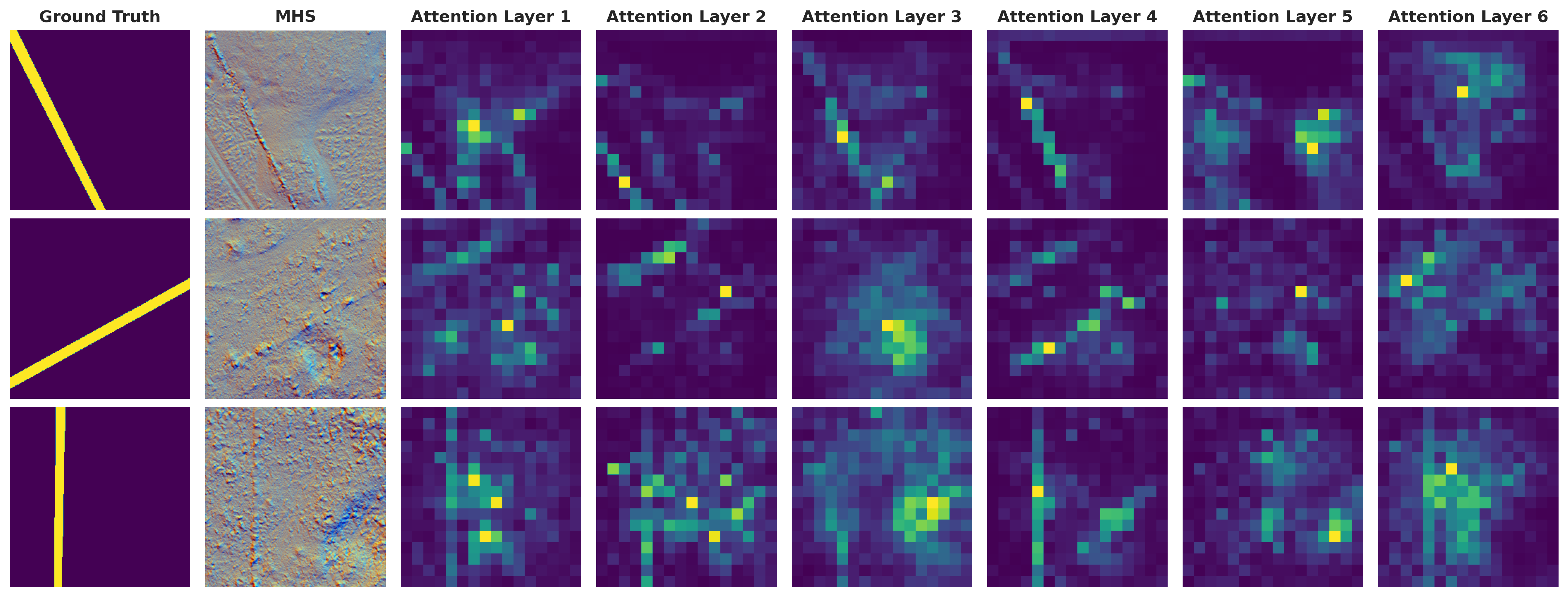}} 
    \hfill
    \subfloat[Attention feature maps on VAT inputs.]{\includegraphics[width=1\textwidth]{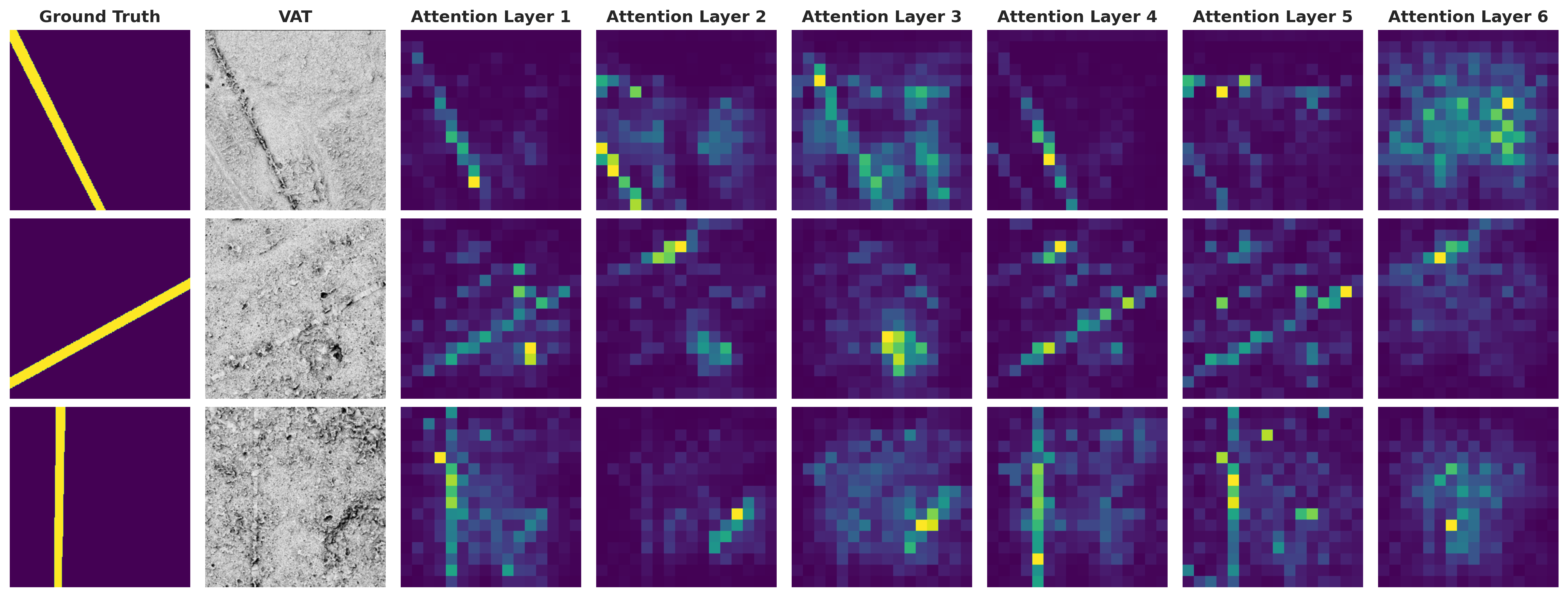}} 
\caption{Self-attention maps from the pre-trained DINO-CV ViT-S/16 model without supervised fine-tuning.}
\label{fig:vit_atten_maps}
\end{figure*}

\begin{figure*}[!tbp]
    \centering
    \subfloat[ViT backbone]{\includegraphics[width=0.5\textwidth]{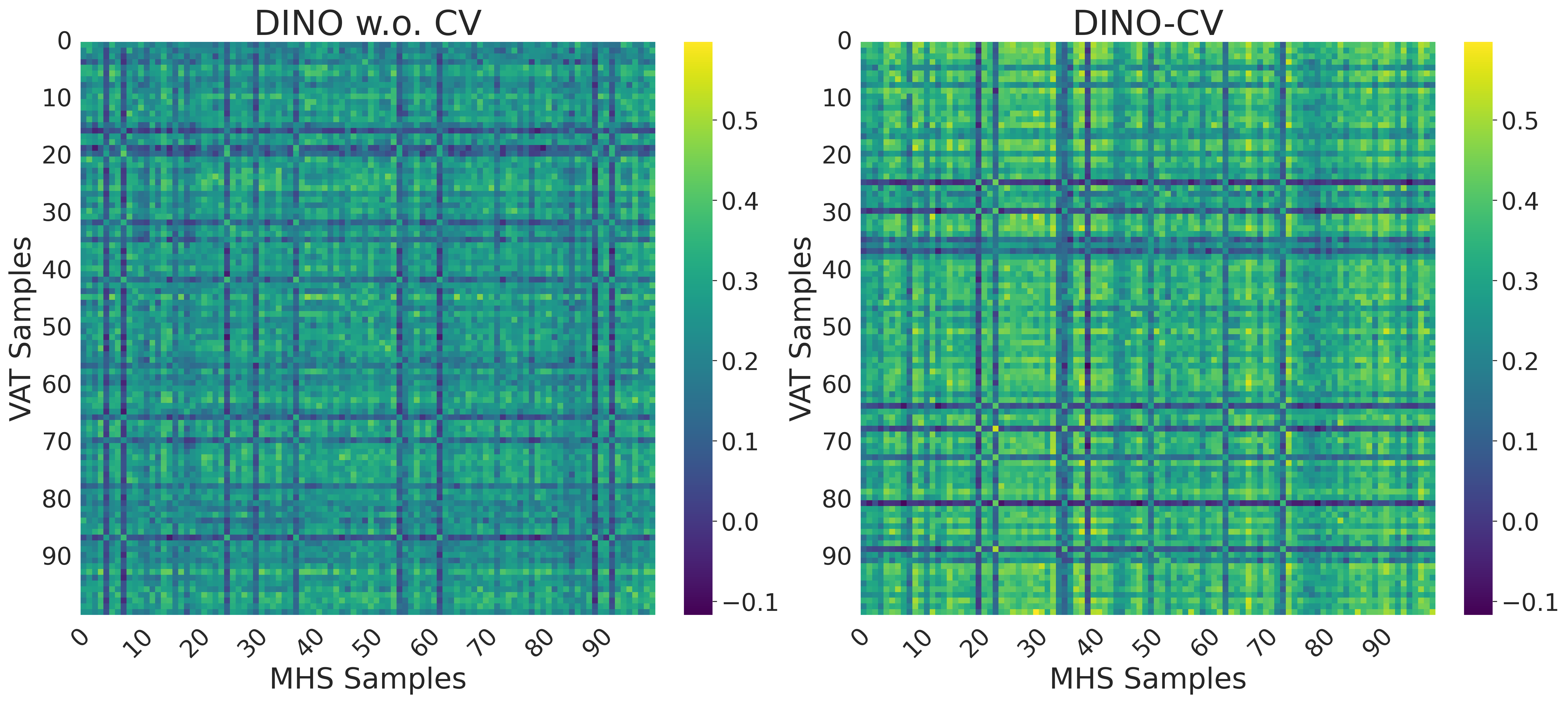}} 
    \hfill
    \subfloat[ResNet backbone]{\includegraphics[width=0.5\textwidth]{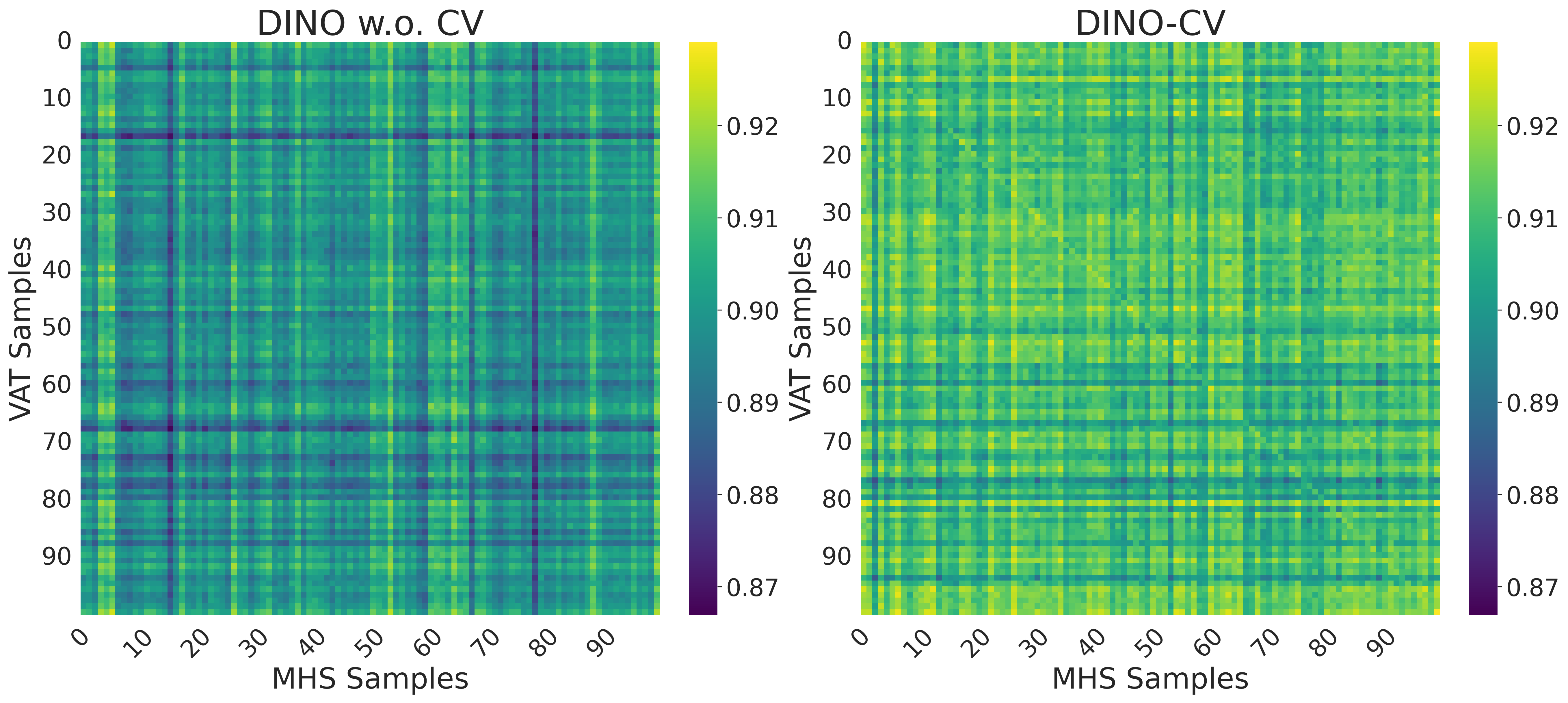}} 
    \caption{Feature similarity maps of ViT and ResNet backbones on MHS and VAT inputs. MHS and VAT samples are randomly selected from the BudjBimArea dataset. The diagonal of each matrix indicates the cosine similarity between MHS and VAT features from the same location, reflecting the degree of cross-view alignment.}
    \label{fig:dino_cv_feat_sim}
\end{figure*}

\paragraph{Ablation Study on DINO-CV}
In Table~\ref{tab:ablation}, we conduct an ablation study to evaluate the effectiveness of two key design choices of our mapping approach: (1) the use of DINO-CV for pre-training, and (2) the Pseudo-Siamese architecture for fusing representations from multiple DEM derivatives (MHS and VAT) during supervised fine-tuning. 

Across all backbones (i.e., ResNet-50, WRN-50-2, and ViT-S/16), removing the cross-view pre-training component (\textcolor{gray}{rows 2, 6, 10}) leads to consistent performance drops in both F1 and mIoU, confirming the effectiveness of cross-view pre-training in learning view-invariant representations. This finding supports our core hypothesis that enforcing alignment across complementary DEM modalities enhances the model capability for mapping. Furthermore, replacing the dual-stream Pseudo-Siamese architecture with a single-stream architecture (either MHS or VAT; \textcolor{gray}{rows 3–4, 7–8, 11–12}) also results in lower performance. The improvements brought by the dual-stream data view fusion demonstrate the complementary nature of MHS and VAT features, validating the Pseudo-Siamese architectural design. Together, these results show that both the cross-view pre-training and the dual-stream Pseudo-Siamese architecture contribute positively and independently to model performance, with the full DINO-CV configuration achieving the best results across all backbones.

\paragraph{Learning View-invariant Features} 
To gain insight into the spatial attention patterns of models trained with DINO-CV, we visualize the self-attention maps produced by the ViT-S/16 backbone in Fig.~\ref{fig:vit_atten_maps}, without any supervised fine-tuning. These maps highlight the regions of the input DEM derivatives (i.e., MHS and VAT) that the model attends to during inference. The visualizations reveal that DINO-CV enables the model to consistently focus on linear and associated geomorphic features resembling dry-stone walls. In particular, attention heads in intermediate ViT layers (i.e., attention layers 4 and 5) emphasize wall-like structures, while other attention layers (i.e., attention layer 1, 3 and 6) capture broader contextual information, such as the surrounding terrain and slope. 

To further evaluate the alignment of feature representations across data views, we analyze the cosine similarity between the latent features extracted from MHS and VAT inputs using the teacher networks of DINO-CV. This analysis is conducted on both ViT and ResNet backbones, without any supervised fine-tuning. As shown in Fig.~\ref{fig:dino_cv_feat_sim}, DINO-CV exhibits substantially higher cosine similarity along the diagonal of the similarity matrix of data views compared to the original DINO (without cross-view learning), indicating improved feature alignment and enhanced view-invariance. These results confirm that DINO-CV effectively reduces inter-view feature discrepancies, particularly in mid-level spatial regions, and support the hypothesis that cross-view pre-training enables robust and consistent representations across diverse DEM derivatives.

\paragraph{Fine-tuning with Fewer Labels}  
Table~\ref{tab:subset} presents the mIoU performance of various models trained on different proportions (50\%, 30\%, and 10\%) of annotated data from the \textit{BudjBimArea} dataset. This scenario reflects real-world constraints, where annotation is costly and requires domain expertise, particularly in archaeological and heritage landscapes.

Across all backbone architectures, models pre-trained with DINO-CV consistently outperform both randomly initialized and ImageNet-supervised baselines, with the largest gains observed under low-label regimes. Using only 10\% of annotated data, the DINO-CV pre-trained ResNet-50-Siamese model achieves 63.6\% mIoU, compared to 49.3\% for the supervised baseline and 21.4\% for random initialization. A similar pattern is observed for the WRN-50-2-Siamese model, which achieves 63.8\% mIoU at 10\% supervision, outperforming the supervised and randomly initialized baselines by 14.1\% and 41.3\%, respectively.

The largest relative improvement is observed with the ViT-S/16-Siamese model. Under 10\% supervision, the randomly initialized ViT collapses to 14.3\% mIoU, while DINO-CV enables it to retain 58.9\% mIoU, a gain of over 44\% mIoU. Even compared to the supervised ViT (56.4\%), DINO-CV improves performance across all levels of label availability.

These results highlight the value of cross-view self-supervised learning in enabling data-efficient feature learning for object mapping in archaeological landscapes. By leveraging unlabeled DEM derivatives, DINO-CV significantly reduces the reliance on large annotated datasets while maintaining high segmentation performance across both ConvNet and ViT backbones. This makes our approach especially suitable for large-scale, label-scarce tasks such as dry-stone wall mapping in cultural heritage environments.

\section{Conclusions}
\label{conclusion}
In this study, we introduced \textbf{DINO-CV}, a cross-view self-supervised learning framework for mapping dry-stone walls in archaeologically significant and visually complex landscapes. Leveraging multiple DEM-derived views of archaeological landscape, MHS and VAT, DINO-CV enables models to learn view-invariant and semantically rich features in self-supervised pre-training. Our method significantly improves stone wall mapping performance over both supervised and existing self-supervised baselines across ConvNet and ViT backbones. Under limited annotation, DINO-CV demonstrates strong label efficiency and intra-site generalization, making it highly suitable for large-scale cultural heritage mapping tasks.

\paragraph{Limitations}
An inevitable limitation lies in the accuracy of wall annotations in the \textit{BudjBimArea} dataset. Given the challenges in visually interpreting dense vegetation, shadowing, and subtle terrain features, some wall labels may suffer from spatial uncertainty or omission. Additionally, while the dataset focuses on a geographically and culturally rich region, its domain specificity may limit the direct transferability of results to other archaeological landscapes with different geomorphic characteristics or wall typologies.

\paragraph{Future Work}
Future work could further explore improving the annotation quality of the \textit{BudjBimArea} dataset through semi-automated labeling strategies and integrating expert feedback loops. Additionally, extending self-supervised pre-training to cross-platform transfer learning on LiDAR data (UAV-airborne LiDAR) represents a promising research direction. For instance, while UAV-LiDAR enables fine-grained capture of low-relief archaeological features at local scales, acquiring reliable annotations remains challenging and resource-intensive. The cross-view pre-training strategy of DINO-CV, designed to extract invariant geometric features from multiple DEM derivatives, is theoretically well suited to bridge the domain gap between regional airborne LiDAR surveys and high-resolution UAV-LiDAR acquisitions. Transferring models pre-trained on extensive airborne datasets to UAV-based surveys could therefore mitigate annotation scarcity and enable high-resolution mapping without requiring extensive new labels. Finally, applying DINO-CV to additional cultural heritage landscapes worldwide would help further assess its generalizability and scalability across diverse geomorphic settings and stone wall typologies.

\section*{CRediT authorship contribution statement}
\textbf{Zexian Huang}: Conceptualization, Writing – original draft, Writing – review \& editing, Software, Methodology, Investigation. \textbf{Mashnoon Islam}: Conceptualization, Data curation. \textbf{Brian Armstrong}: Writing – review \& editing. \textbf{Billy Bell}: Project administration, Supervision. \textbf{Kourosh Khoshelham}: Writing – review \& editing, Supervision. \textbf{Martin Tomko}: Writing – review \& editing, Supervision.

\section*{Declaration of Competing Interest}
The authors declare that they have no known competing financial interests or personal relationships that could have appeared to influence the work reported in this paper.

\section*{Funding sources}
This research was supported by the Australian Research Council (ARC) under grant SR200200227. The authors gratefully acknowledge the Department of Environment, Land, Water and Planning (DELWP), Victoria, for sourcing and providing access to the airborne LiDAR dataset. All data and derived results are owned by the Gunditj Mirring Traditional Owners Corporation (GMTOC), with research conducted in collaboration under an established memorandum of understanding (MoU).

\section*{Declaration of generative AI use}
During the preparation of this work, the authors used \textit{ChatGPT} (OpenAI, San Francisco, USA) to assist in improving language clarity, grammar, and readability of the manuscript. All content was carefully reviewed, verified, and edited by the authors to ensure accuracy, coherence, and that the final text reflects their original research, interpretation, and conclusions. The authors take full responsibility for the content of the published article.

\section*{Data availability}
\label{availability}
The source code and data supporting the findings of this study are available at: \url{https://github.com/MLinArcheaomatics/BudjBimStoneWall-DINO-CV}.

\section*{Acknowledgements}
\label{acknowledgements}
We thank Gunditj Mirring Traditional Owners Corporation (GMTOC) for providing the research data and overseeing its use in this study. Their invaluable cooperation, support, and meticulous efforts significantly contributed to the quality of this research.


\bibliographystyle{elsarticle-harv} 
\bibliography{cas-refs}

\end{document}